\documentclass{article}

\usepackage{arxiv}

\usepackage[utf8]{inputenc}
\usepackage[T1]{fontenc}
\usepackage{hyperref}  
\usepackage{url}        
\usepackage{booktabs}   
\usepackage{amsfonts}   
\usepackage{amsmath}
\usepackage{nicefrac}   
\usepackage{microtype}  
\usepackage{lipsum}
\usepackage{graphicx}
\usepackage{longtable}
\usepackage{array}
\usepackage{ragged2e}

\graphicspath{ {./images/} }

\title{Language of Persuasion and Misrepresentation in Business Communication: A Textual Detection Approach}

\author{
 Sayem Hossen \\
  Department of Computer Science \& Engineering\\
  Sonargaon University\\
  Dhaka-1215, Bangladesh \\
  \texttt{sayemhossen.contact@gmail.com} \\
   \And
  Monalisa Moon Joti \\
  Department of Computer Science \& Engineering\\
  Sonargaon University\\
  Dhaka-1215, Bangladesh \\
  \texttt{monalisamoonsapp@gmail.com} \\
    \And
  Md. Golam Rashed \\
  Dept. of Information and Communication Engineering\\ Faculty of Engineering\\
  University of Rajshahi\\
  Rajshahi-6205, Bangladesh \\
  \texttt{golamrashed@ru.ac.bd} \\
}

\begin{document}
\maketitle
\begin{abstract}
Business communication digitisation has reorganised the process of persuasive discourse, which allows not only greater transparency but also advanced deception. This inquiry synthesises classical rhetoric and communication psychology with linguistic theory and empirical studies in the financial reporting, sustainability discourse, and digital marketing to explain how deceptive language can be systematically detected using persuasive lexicon. In controlled settings, detection accuracies of greater than 99\% were achieved by using computational textual analysis as well as personalised transformer models. However, reproducing this performance in multilingual settings is also problematic--and, to a large extent, this is because it is not easy to find sufficient data, and because few multilingual text-processing infrastructures are in place. This evidence shows that there has been an increasing gap between the theoretical representations of communication and those empirically approximated, and therefore, there is a need to have strong automatic text-identification systems where AI-based discourse is becoming more realistic in communicating with humans.
\end{abstract}


\section{Introduction}
The transparency and authenticity of organisations have changed due to digitalisation in corporate practice. Although the technological advancements have brought a new life with unparalleled opportunities, the same expose organisations to vulnerabilities that are detrimental, and the most eminent one is the ability of persuasive language to be beckoned and directed to explicit lying. Such weakness will be enhanced even further when the parties involved share written materials instead of interacting directly when they come to the point of making decisions with considerable impacts (Zhou \& Zhang, 2008)\cite{zhou2008statistical}. New empirical findings indicate that artificial intelligence, being a dual-use technology, has acquired advanced tools of fraud in addition to the advanced mechanism of fraud detection, thus establishing a multifaceted technological and ethical environment whereby working practices should be enhanced constantly (Park et al., 2024)\cite{park2024ai}.

Through the merger of the high standards of computational methods that can analyse large corpora with a significant level of reliability and speed (Bochkay et al., 2023)\cite{bochkay2023textual}, the current empirical investigation contributes to the previous research in rhetorical analysis extensively. The negative consequences of unregulated veracity are reflected in the financial consequences of false financial reporting, both in local and foreign jurisdictions, which have suffered losses in thousands of dollars (Humpherys et al., 2011)\cite{humpherys2011identification}, and in environmental effects of greenwashing initiatives through the proxy interrogatives of corporate climate engagement (Caputo et al., 2021)\cite{caputo2021enhancing}. Most importantly, the world witnessed a 35\% greenwashing growth in 2023, with the most abhorrent miscreants being in the oil-and-gas and financial spheres (Forliano et al., 2025)\cite{forliano2025mapping}. When combined, each of these findings raises the need to consider highly effective detection solutions presumably rooted in highly specified machine-learning models as AI increasingly develops the ability to create highly realistic, human-sounding language that will excite even experienced detectives (Shah et al., 2023)\cite{shah2023detecting}.

The current literature review summarises the literature devoted to finding persuasion and misrepresentation in texts. The main contributions that can be highlighted are as follows: \textbf{(1)} contrary to the advancements made in distinct areas (predominantly financial reporting and online reviews), a universal methodology to analyse the business communication landscape as a single entity, has not been identified; \textbf{(2)} communication studies, the field of psychology, linguistics, and computational detection intersect; this avenue could lead to more discrete and fruitful lines of enquiry; and \textbf{(3)} the notion of an English first prejudice, along with the proliferation of multilingual applications, is an important part of resolving the current methodological limits.

The paper is organized as follows: In Section 2, the literature review of the already available literature on the theme of persuasion and misrepresentation in business communication will be presented. Section 3 will describe the methodology of this research, namely, how the datasets were acquired, pre-processed, and engineered and the architecture of the models. Section 4 shows and discusses the results of the experiments. The section 5 is about the theoretical and practical implications of the findings. Lastly, Section 6 provides the limitations, the future research directions, and concluding thoughts that conclude the paper.

\section{Literature Review}
The literature that is available on the subject of persuasion and misrepresentation in business communication brings out the complexity of recognizing and deciphering misleading patterns in business communication. The examples of the application of language selection, rhetoric and psychological strategies to persuade perceptions and direct behaviours are given in various studies. The most significant theoretical models, such as the Elaboration Likelihood Model (ELM) and the framing theory, are mentioned in detail as regards to the persuasion dynamics. Further, there are linguistic cues, including the use of pronouns, metadiscourse, and framing structures, which are revealed to be critical in deceptive activity. The new advances in computational techniques, especially the machine learning (ML) and deep learning systems have enhanced the ability to identify and categorize such deceptive practices, especially in the financial reporting, sustainability communications, and online marketing scenarios.

{\small
\renewcommand{\arraystretch}{1.3}
\begin{longtable}{
  >{\raggedright\arraybackslash}p{3.5cm}
  >{\raggedright\arraybackslash}p{3.3cm}
  >{\raggedright\arraybackslash}p{6.7cm}
}

\caption{Summary of Background Study} 
\label{tab:studies}\\

\toprule
\textbf{Author(s)}  & \textbf{Method} & \textbf{Contribution} \\
\midrule
\endfirsthead

\multicolumn{3}{c}%
{{\tablename\ \thetable{} -- continued from previous page}} \\
\toprule
\textbf{Author(s)} & \textbf{Method} & \textbf{Contribution} \\
\midrule
\endhead

\bottomrule
\multicolumn{3}{r}{{Continued on next page}} \\
\endfoot

\bottomrule
\endlastfoot

Alkaraan et al., 2023\cite{alkaraan2023} & Rhetoric-based analysis & Found marketing campaign slogans that produce false hope and cover the financial risks. \\[3pt]
Camilleri, 2022\cite{camilleri2022walking} & Elaboration Likelihood Model & Demonstrated the effect of information quality on the discourse of CSR and the attitude of stakeholders. \\[3pt]
Flusberg et al., 2024\cite{flusberg2024psychology} & Framing theory & Reviewed the mechanism of how framing influences thinking and behavior, and revealed how language is misused. \\[2pt]
Dimmelmeier, 2021\cite{dimmelmeier2021sustainable} & Framing analysis & Investigated the impacts of repetitive patterns in framing on how people view sustainability as an artefact of power arrangements. \\[2pt]
Malyuga, 2023\cite{malyuga2023corpus} & Corpus-based linguistic analysis & Illustrated how personal pronouns are employed by corporations to create pseudocollectivism among the stakeholders. \\[2pt]
Liu \& Zhang, 2021\cite{liu2021using} & Metadiscourse analysis & Discovered that corporate press release is more dependent on attitude markers than objective transmission of information. \\[2pt]
Scorici et al., 2024\cite{scorici2024anthropomorphization} & Linguistic analysis & The hyperbolic statements about the ability of AI by companies became known as humanwashing. \\[2pt]
Pizzi et al., 2021\cite{pizzi2021voluntary} & Cultural analysis of CSR & Established the effect that cultural dimensions have on successful persuasion and detecting deception accuracy. \\[2pt]
Yang et al., 2023\cite{yang2023human} & Influencer credibility & Demonstrated that perceived humanness of social media influencers can influence the credibility of the CSR messages. \\[2pt]
Kumar, 2023\cite{kumar2023ethical} & Psychological targeting & Studied the nature of AI systems and how they can control consumer behavior through the implementation of psychological targeting, which is not ethical. \\[2pt]
Park et al., 2025\cite{park2024ai} & AI deception analysis & Was able to put forward how AI can possibly takes misleading tricks in training unknowingly. \\[2pt]
Yang et al., 2023\cite{yang2023finchain} & FinChain-BERT model & FinChain-BERT, the suggested method of analyzing financial fraud using the context in examination. \\[2pt]
Caputo et al., 2021\cite{caputo2021enhancing} & Greenwashing analysis & Discovered that the companies withhold, albeit strategically, negative environmental news in order to save brand repute. \\[2pt]
De Villiers et al., 2023\cite{devilliers2023will} & AI-generated content & Researched how AI creates duplicates of the current ones of greenwashing and continues the misrepresentation. \\[2pt]
Oppong-Tawiah \& Webster, 2023\cite{oppong2023} & Linguistic detection & 
Discovered that there was a severe negative relationship between greenwashing language and financial performance. \\[2pt]
Bello et al., 2023\cite{bello2023machine} & Machine learning & Detecting the fraud of financial transactions based on finding the patterns of language using Applied ML. \\[2pt]
Javed et al., 2021\cite{javed2021fake} & CNN model & Applied CNN to fake review detection and has been capable of showing high F1 scores. \\[2pt]
Nabrawi \& Alanazi, 2023\cite{nabrawi2023fraud} & Random forest & The accuracy achieved with regard to detecting healthcare insurance fraud was 98.21\% \\[2pt]
Bahoo et al., 2024\cite{bahoo2024artificial} & ML comparison & SVMs, neural networks, and Random Forests are very useful in identifying financial fraud. \\[2pt]
Aslam et al., 2022\cite{aslam2022interpretable} & Explainable AI & Proposed an XAI system, whereupon identification of malicious domains, it can be ascertained that it is 98.56 accurate. \\[2pt]
Wang \& Chen, 2023\cite{wang2023attentive} & Deep learning & Applied a mix of written and quantitative features to come to high precision decisions when it came to detecting financial fraud. \\[2pt]
Abayomi-Alli et al., 2022\cite{abayomi2022deep} & Bi-LSTM & Reported 93.4\% and 98.6\% spam detection of SMS. \\[2pt]
Oswald et al.,2022\cite{oswald2022spotspam} & BERT model & Obtained 98.07\% accuracy to detect the SMS spam with BERT. \\[2pt]
Hashmi et al., 2024\cite{hashmi2024advancing} & Hybrid deep learning & Proposed a mixed model of the fake news detection with the accuracy of 99\%. \\[2pt]
Zhu et al., 2021\cite{zhu2021intelligent} & Graph neural networks & Was aware that GNNs were trendy to spot post-COVID financial fraud. \\[2pt]
Elluri et al., 2021\cite{elluri2021bert} & BERT for compliance & Provided a BERT-based GDPR compliance model that was able to give 75\% accuracy. \\[2pt]
Gupta, 2023\cite{gupta2023gpt} & GPT-InvestAR & Applied used GPT to forecast financials out of annual reports. \\[2pt]
Li et al., 2023\cite{li2023extracting} & LLM data extraction & Reached 99.5\% of automating extraction of financial information contained in PDFs. \\[2pt]
Hassani et al., 2024\cite{hassani2024rethinking} & LLM compliance & The GPT-4 as a model of automation has been tested to be in legal compliance with an accuracy of 40 percent. \\[2pt]
Qi et al., 2024\cite{qi2024sniffer} & Multimodal LLM & Trained a multimodal LLM capable of identifying misinformation and could beat GPT-4V by 11 percent in accuracy. \\[2pt]
Mayank et al., 2022\cite{mayank2022deap} & Knowledge graph & Used NLP and tensor decomposition to accomplish the problem of fake news detection with a high precision. \\[2pt]
Hajek et al., 2022\cite{hajek2022fake} & Sentiment analysis & Beat conventional sentiment analysis on the aspect of detecting fake reviews. \\[2pt]
Wang et al., 2023\cite{wang2023attentive} & RCMA model & Applied a ratio-based model to assess financial statements and calculated AUC equal to 86.2 per cent. \\[2pt]
Chen et al., 2021\cite{chen2021opinion} & Opinion mining & Argumentative structures of reports were analyzed in order to formulate financial opinion mining. \\[2pt]
Kang \& Kim, 2022\cite{kang2022} & NLP analysis & Determined sentence similarity to be a better alternative to keyword matching when it comes to finding corporate self-presentation. \\[2pt]
\bottomrule
\end{longtable}
}

The existing literature review presented in Table ~\ref{tab:studies} is a combination of the available literature on the inter-relationships between the theories of communication, linguistic markers, and contemporary computational techniques that may be adopted to identify misrepresentation and persuasion in business communication. As the articles reviewed show, linguistic characteristics such as framing, metadiscourse, and personal pronouns play a crucial role in the corporate discourse. They can be effectively utilized with the aim of deception. The complementary effect of conventional communication paradigms and the most recent AI technologies, particularly machine and deep learning, can provide novel methodological solutions to the emerging necessity of identifying deception in various corporate environments. The next step of this study is further refinement of these analytical models, accounting for cross-cultural and multilingual complexity and making AI-based approaches explainable.

\section{Methodology}
This study presents a comprehensive computational framework for detecting persuasive and misleading language patterns in business communication, integrating advanced deep learning architectures with rigorous text preprocessing and evaluation protocols. Our methodology addresses the linguistic complexities of business discourse through a multi-stage pipeline encompassing data collection, preprocessing, feature engineering, model development, and performance evaluation. The approach is grounded in established natural language processing paradigms for deception detection while incorporating innovations specifically tailored to business communication contexts \cite{zhou2008statistical, humpherys2011identification}.

\subsection{Dataset Collection and Preprocessing}
The current study is a complete work on the basis of an especially prepared corpus, i.e., 4,848 instances of business communication,  as detailed in Table~\ref{tab:dataset_distribution} that were fully annotated by three business communication researchers, a linguist, and a computational linguist. The information was drawn from 13 heterogeneous modalities that include marketing emails, social-media advertising on Twitter/X Facebook, financial statements, lies in corporate statements, newspaper articles, LinkedIn postings, YouTube advertising, international-business reports, and marketing-criteria reports. Such a multi-source architecture gives access to a large set of potentially convincing and/or deceitful speech artefacts, a gap in coverage that previous studies have found in more generally applicable deception detection systems (Baden et al., 2022)\cite{baden2022three}.

The resulting dataset is made up of 1,980 Factual, 30.5 (mean: Persuasive contents, and 28.7 Misleading communications, hence having a balanced distribution across classes, which cleans the bias generated by the imbalanced nature of conventional deception-detection tasks (Glockner, 2022)\cite{glockner2022missing}. Two experts annotated each text, and any disparity between the two experts was systematically resolved in agreement or through reference to the decision of a third expert. Annotation schemes were harmonised with the existing models of marking deceitful discourse (Zhou, 2008; Humpherys, 2011)\cite{zhou2008statistical, humpherys2011identification} and modelled on the examples of indicators provided in previous studies (Larcker, 2012; Craig, 2013)\cite{larcker2012detecting,craig2013exploring}, such as mention of uncertainty, extremely positive affect, non-immediacy, and framing.

\begin{table}[h]
\centering
\caption{Dataset Distribution}
\begin{tabular}{lcc}
\toprule
Class & Instances & Percentage \\
\midrule
Factual & 1,980 & 40.8\% \\
Persuasive & 1,479 & 30.5\% \\
Misleading & 1,389 & 28.7\% \\
\midrule
Total & 4,848 & 100\% \\
\bottomrule
\end{tabular}
\label{tab:dataset_distribution}
\end{table}

\begin{figure}[h]
\centering
\includegraphics[width=0.8\textwidth]{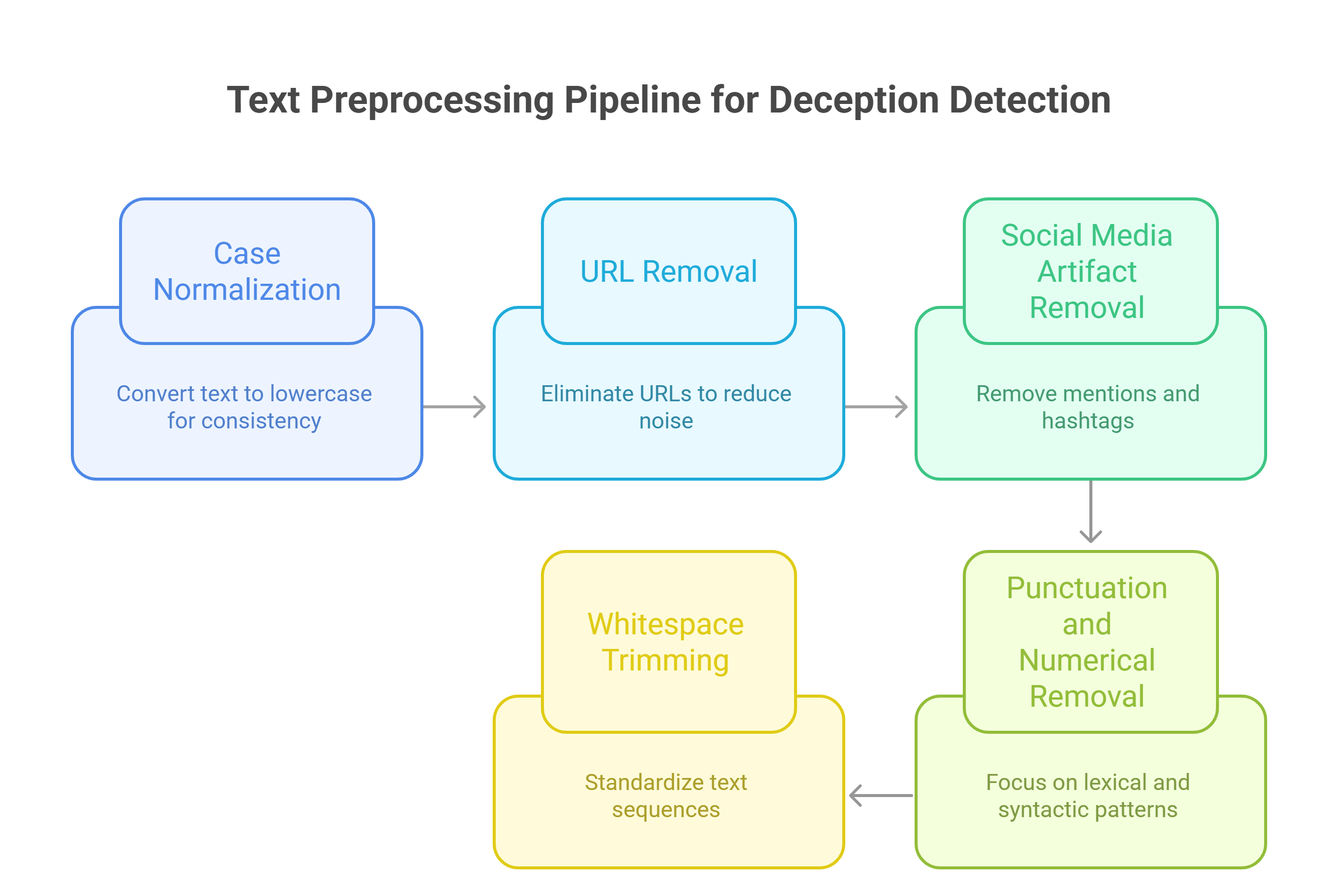}
\caption{Five-step text preprocessing pipeline for deception detection}
\label{fig:preprocessing_pipeline}
\end{figure}

A well-designed and structured five-step pipeline was implemented on the processing of textual data pertaining to business. Both stages were aimed at achieving consistency in linguistic properties of the selected corpus as well as ensuring the retention of prominent semantic material required to detect instances of deception. The initial step involved the process of case normalisation, i.e., all characters were transcribed to lower case thus improving the balance of lexical consistency and reducing variance and homonymy of similar lexical items. The next step consisted of the removal of syntactic classes like embedded links which were eliminated through the use regularly occurring expressions (http\S+, www\S+, https\S+). Such a process was based on a previous empirical work that indicated such factors often represent extraneous noise in deception-detection exercises  \cite{zhou2004comparison}, and its elimination, therefore, limits more concise semantic modeling. At the third stage, social-media artefacts, i.e. mentions @username and hashtags stipulated with a symbol hash, were systematically eliminated so as to reduce platform-specific noise and artifactual effects that could otherwise interfere with semantic analysis. Stage four corresponded to the warranted extraction of punctuation marks and the digits followed by number values, with respect to the practices that are traded within the literature of financial-fraud detection whereby focus is centred on the fundamental units of the lexicon and syntax structure rather than the number tokens. Lastly, the stage five would be used to correct the residual formatting problems by removing the extra whitespaces leaving fully normalised, regular textual strings. Taken together, the above preprocessing steps alleviated the curse of dimensionality effect that high dimensionality can have on the integrity of business text and helped preserve the integrity and representational power of the semantic content contained in the communication at the business level. The flow diagram of Figure~\ref{fig:preprocessing_pipeline} can be evidence of it.

\subsection{Feature Engineering and Text Representation}

The current text tries to place the business communication in the framework of deep learning. In this endeavour, a tokenisation-based extraction method was developed to preserve semantic richness and make it scalable at the same time without compromising computation. A Tokeniser was also set up as a Keras Object with a vocabulary limit of 10 000 lexical units and an out-of-vocabulary (OOV) clone. The arrangement strikes the equilibrium between retaining the context-specific discursive aspects of business communication and the demands of computation inherent to specialised domains (Purda \& Skillicorn, 2012)\cite{purda2012accounting}.

After tokenisation, input texts were transformed into an ordered sequence of integers that specifically matched the vocabulary. In order to obtain a homogenous maximum sequence length, every sequence padding or truncation was carried out to 100 tokens- a decision that was informed by post-processing methods (Flusberg et al., 2024)\cite{flusberg2024psychology}. A truncation threshold is equal to several standard observations pointing out that 95\% of all the input texts naturally fall inside of this range, which triggers the speed of the training process by mitigating padding at the sequence level as well as the propagation of loss related to it.

Multi-class label encoding was done by numerically mapping the categorical target classes: Factual, Misleading, and Persuasive, with the \textbf{LabelEncoder} utility of the scikit-learn library. The label codes were Factual=0, Misleading=1, and Persuasive=2. This encoding process is entirely in line with the categorical cross-entropy loss that is used during the training of the model and plays a key role in matching the performance measures with its internal architecture.

\begin{figure}[h]
\centering
\includegraphics[width=0.8\textwidth]{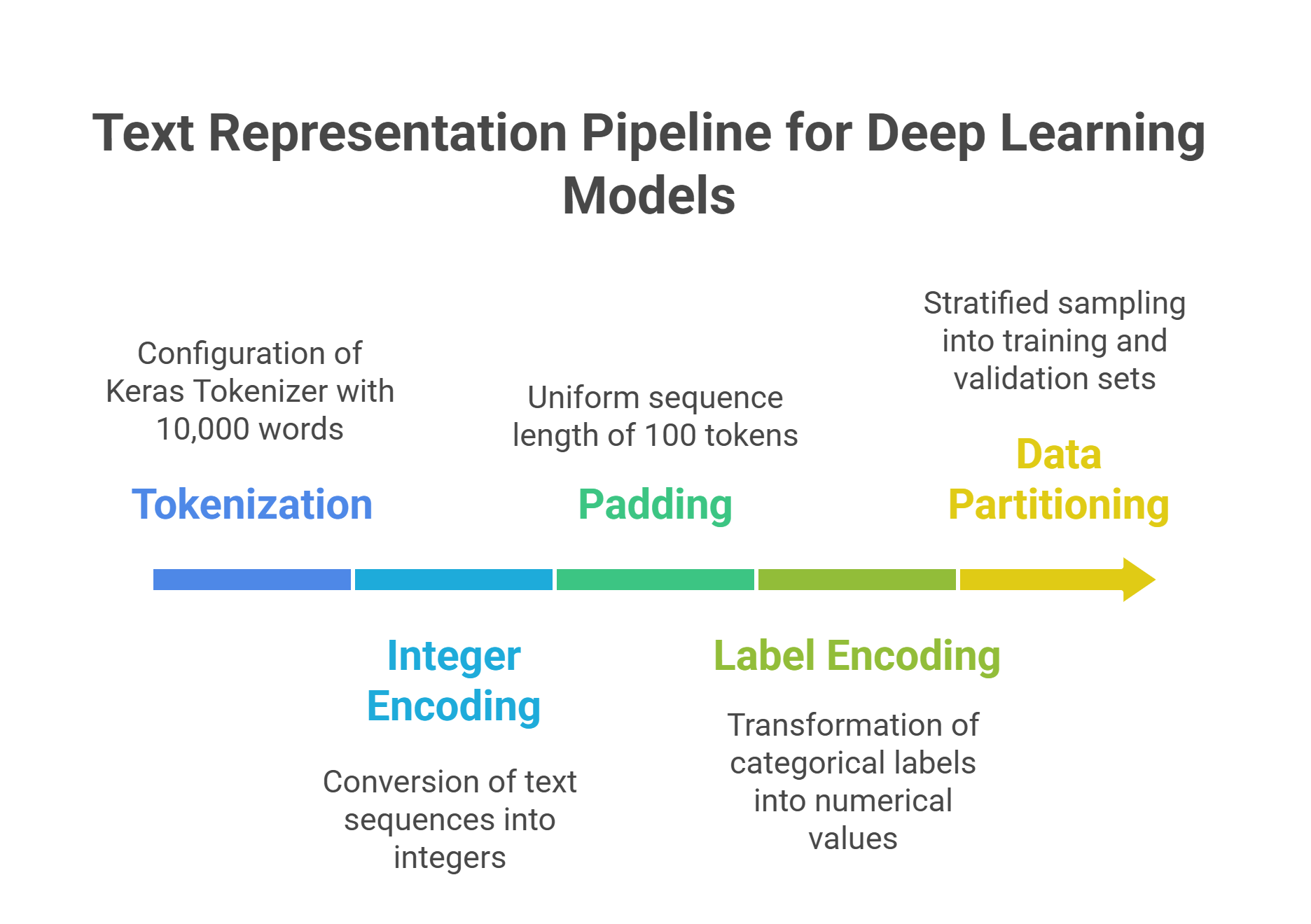}
\caption{Text representation pipeline, from raw text to tokenized sequences and padding for model input.}
\label{fig:text_representation_diagram}
\end{figure}

To ensure that the dataset used should retain the intrinsic class distributions of the original dataset and to create training data which is balanced (i.e. with equal classes) the corpus was divided into two mutually exclusive subsets, a training and validation set, where the former had 3,878 samples (80\% of the total) and the latter 970 samples (20\% of the total). The division was performed using stratified sampling, a published approach to alleviating the problem of overfitting and enhancing generalisability, especially in the comparatively small corpus sizes typical of this task. Figure~\ref{fig:text_representation_diagram} shows an example of an end-to-end pipeline that includes pre-processing and splitting of data.

\subsection{Model Architectures}
In the present study, we benchmarked five different deep-learning architectures to give a complete evaluation of detection capabilities on different computational paradigms, as summarized in Table~\ref{tab:model_comparison}. Chosen based on the fact that they were proven useful in textual analysis and adapted to the business communication analysis of deception detection task, the architecture types are sequential models (Bidirectional LSTM variations), attention-based mechanisms (Custom Attention and Transformer), and convolutional networks (CNN). Such architectural heterogeneity allowed a systematic comparison of the ways that various computational modalities represent deception-revealing linguistic patterns. Both architectures involved domain-specific modifications (the bidirectional processing to maintain information regarding contexts, the attention to emphasise well-known deception cues (e.g. expressions of uncertainty, extreme sentiment) and the multi-scale convolutional kernels to detect n-grams). This multifold approach was adapted to assess the relative significance of temporal dependencies, contextual relationships, and local discriminative features of identifying deceptive business language on the one hand and to set up strong benchmarks of performance and provide theoretical insights into the computational processes of identifying a deceptive business language on the other hand.

\begin{table}[h]
\centering
\caption{Model Architecture Comparison}
\begin{tabular}{lcccc}
\toprule
Model & Parameters & Layers & Key Innovation & Training Time (s/epoch) \\
\midrule
Simple LSTM & 714,499 & 4 & Bidirectional processing & 1--2 \\
Advanced LSTM & 1,424,387 & 5 & Stacked BiLSTM & 2--3 \\
Custom Attention & 3,143,783 & 8 & Attention mechanism & 3--4 \\
Transformer & 3,917,187 & 7 & Self-attention & 18--19 \\
CNN & 1,397,347 & 9 & Multi-scale convolutions & 1--2 \\
\bottomrule
\end{tabular}
\label{tab:model_comparison}
\end{table}

\subsubsection{Simple Bidirectional LSTM}
The baseline model relies on a Bidirectional LSTM (BiLSTM) network processing the sequences in forward and backward directions in order to receive the contextual relations with the previous words and following words. It is this ability that will be needed to detect the misleading frames that can occur across several phrases \cite{humpherys2011identification}. The model architecture can be characterized as follows: there were embedding layer with vocabulary size = 10000, dimension = 64; the bidirectional LSTM layer with dimension = 64; the dense layer with dimension = 64 and the softmax with three units (the output layer). This architecture has 714,499 trainable parameters that make it a simple, but also a powerful baseline to compare against other more advanced models.

\begin{figure}[h]
\centering
\includegraphics[width=0.8\textwidth]{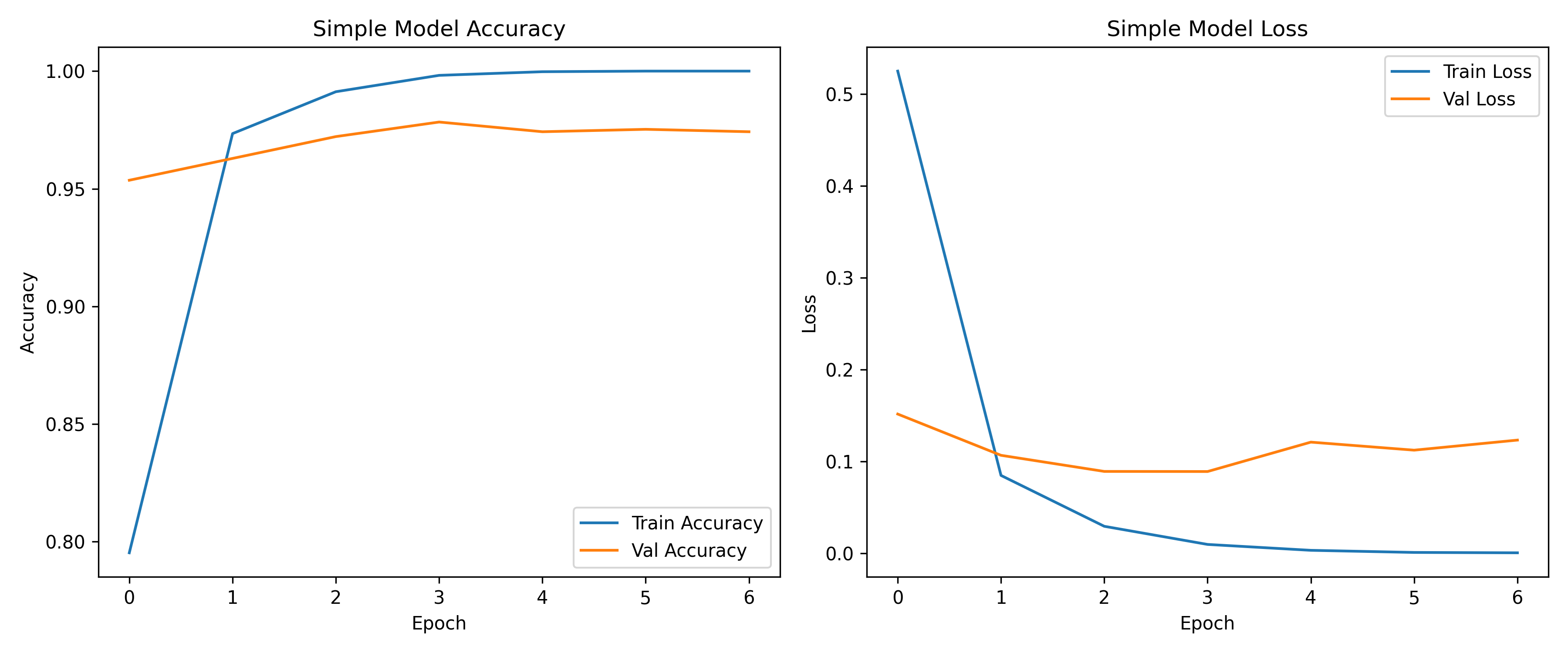}
\caption{Training accuracy and loss curves for the Simple Bidirectional LSTM model.}
\label{fig:simple_model_curves}
\end{figure}

Figure~\ref{fig:simple_model_curves} shows the training dynamics of Simple LSTM model, where training and validation accuracy and loss curves are shown. The accuracy curves (left panel) show that the training accuracy (blue line) reaches sharp increase to approximately 0.8 and nearly to 1.0 at the first 6 cycles and then plateaus, and the validation accuracy (orange line) plateaus at an approximate of 0.975. As seen in the loss curves (right panel), a corresponding rapid drop in training loss toward zero is seen with validation loss slightly increasing after epoch 4 and then stabilizing. The slight difference between the training and validation metrics shows that it might have a good generalization with minimal overfitting.

\subsubsection{Advanced Bidirectional LSTM}
\begin{figure}[h]
\centering
\includegraphics[width=0.8\textwidth]{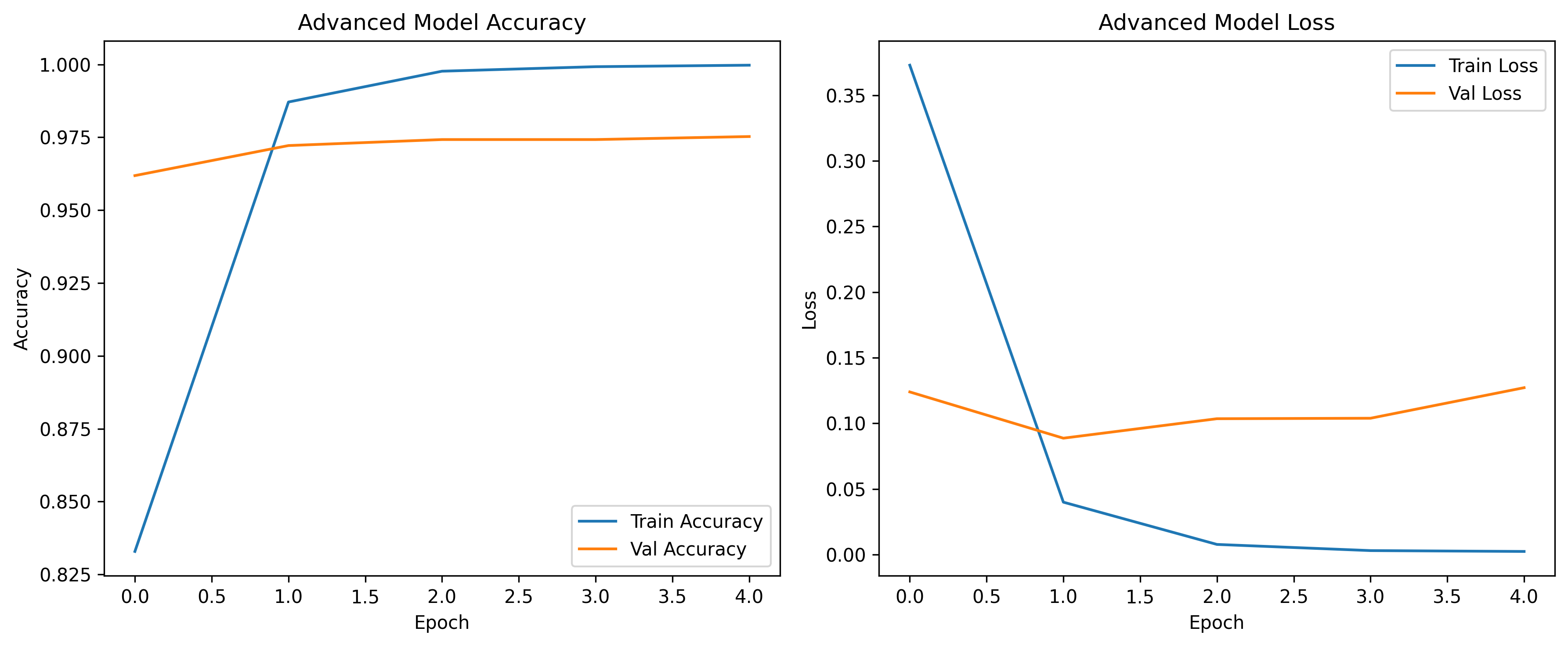}
\caption{Training accuracy and loss curves for the Advanced Bidirectional LSTM model.}
\label{fig:advanced_model_curves}
\end{figure}

The intricate structure augments the baseline by inserting a stacked BiLSTM, hence achieving the hierarchical feature extraction at different levels of linguistic grain and concurrently producing local syntactic patterns and a higher discourse pattern that is relevant to deception detection \cite{ren2016deceptive}. The model structure is as follows: embedding (vocabulary 10,000 words and 128 dimensions), two bidirectional LSTMs (64 and 32 units), respectively, and next the influence of back projections along the sequences, dense (64 units, activation ReLu) and ultimately the output (3 units, activation softmax). This increased capacity enables the description of deceptive linguistic patterns. It maintains a computational efficiency by reducing the dimensions between adjacent layers of LSTM due to the total amount of trainable parameters (1,424,387). As the Figure ~\ref{fig:advanced_model_curves} demonstrates, the training behaviour indicates that while the training accuracy (blue) reaches almost perfect scores (1.0) after the fourth epoch, the validation accuracy (orange) reaches a plateau at roughly 0.97. As the complementary loss plots indicate, training loss (blue) decreases extremely rapidly towards zero, and validation loss (orange) becomes extremely erratic after epoch 2, which is a sign of possible overfitting and non-generalisation as compared to the simpler architecture.

\begin{figure}[h]
\centering
\includegraphics[width=0.8\textwidth]{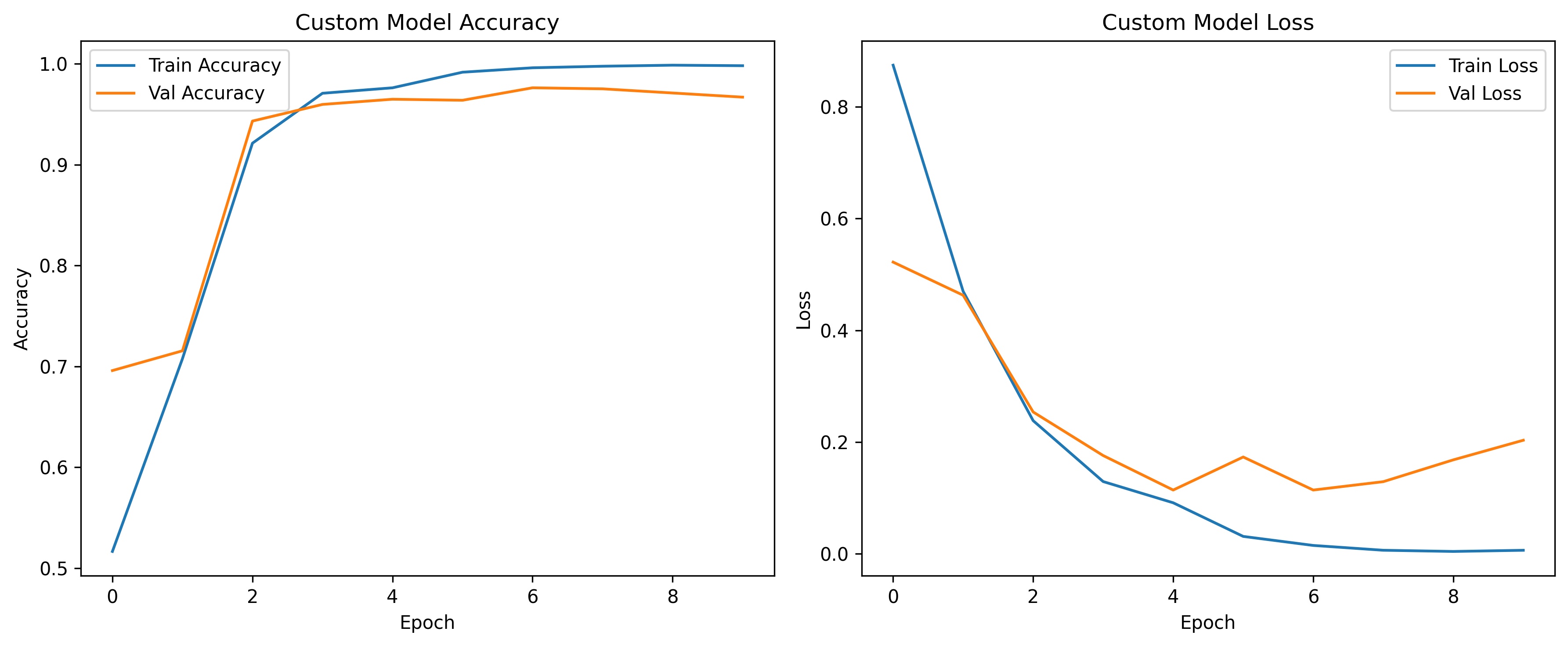}
\caption{Training accuracy and loss curves for the Custom Attention-Enhanced model.}
\label{fig:custom_model_curves}
\end{figure}

\subsubsection{Custom Attention-Enhanced Model}
The offered research outlines a tailor-made architecture incorporating an attention layer, which is capable of clearly modelling the relative significance of various linguistic aspects in deception identification by combining the attention mechanism with the back and forward processing. The module of attention calculates context-conditioned weights at each time step so that the model can focus on the information that is characteristic of deception detected in earlier inquiries, namely, uncertainty expressions, excessive positive emotion, and non-imminent expressions\cite{larcker2012detecting, craig2013exploring}.

The architecture of the created model consists of embedding (which has a size of 10 000 vocabulary size and dimension 256), two stacked bidirectional LSTM layers (256 and 64, sequence-to-sequence layer), a dropout layer (30 percent chance), an implementation of the custom attention layer, two dense layers (128 and 64 units, activated by ReLU), and output (3 units, softmax activation). The architecture (with 3,143,783 trainable parameters) perfectly combines a high level of classification and, at the same time, makes the most interpretable conclusions, highlighting the important parts of the text that will help predict deception.

An example of the progression of training of such a Custom Attention model is depicted in Figure~\ref{fig:custom_model_curves}. In the accuracy (left panel) curves, the learning path is more gradual compared to the observed ones in previous models, with the training accuracy (blue) continuing to advance towards around 0.99 at epoch six and the validation accuracy (orange) gathering maximum accuracy of 0.976 at epoch 7. The loss curves (right panel) show that training loss (blue) is more or less steadily dropping, though validation loss (orange) starts shaky, before making a solid landing on session 5. The existence of the attention mechanism seems to enable more stable convergence patterns, and the narrower distance between training and validation metrics indicates a better generalisation performance as compared to the Advanced LSTM architecture.

\subsubsection{Transformer-Based Architecture}
The use of transformer architectures has been exposed to utilise the self-attention mechanism that highly encourages the execution of diverse rules and regulations of natural language processing \cite{sucholutsky2019attention}. This model is of special suitability in extracting contextual patterns in the discourse of business as it models long-range linguistic relationships. The current experiment uses a variant of the Transformer, which has an embedding layer (vocabulary size 10 000; dimension 256), a custom transformer block (4 attention heads; 512 feed-forward dimension), global average pooling, two dense layers (128 and 64 units; ReLU activation) including dropout (30 percentage rate) as well as an output layer (3 units; softmax activation). Wielding 3,917,187 trainable parameters, the model is the most complex architecture of those compared. It is built to provide a model that can see finer linguistic patterns that might otherwise be lost in a simpler sequential structure.

\begin{figure}[h]
\centering
\includegraphics[width=0.8\textwidth]{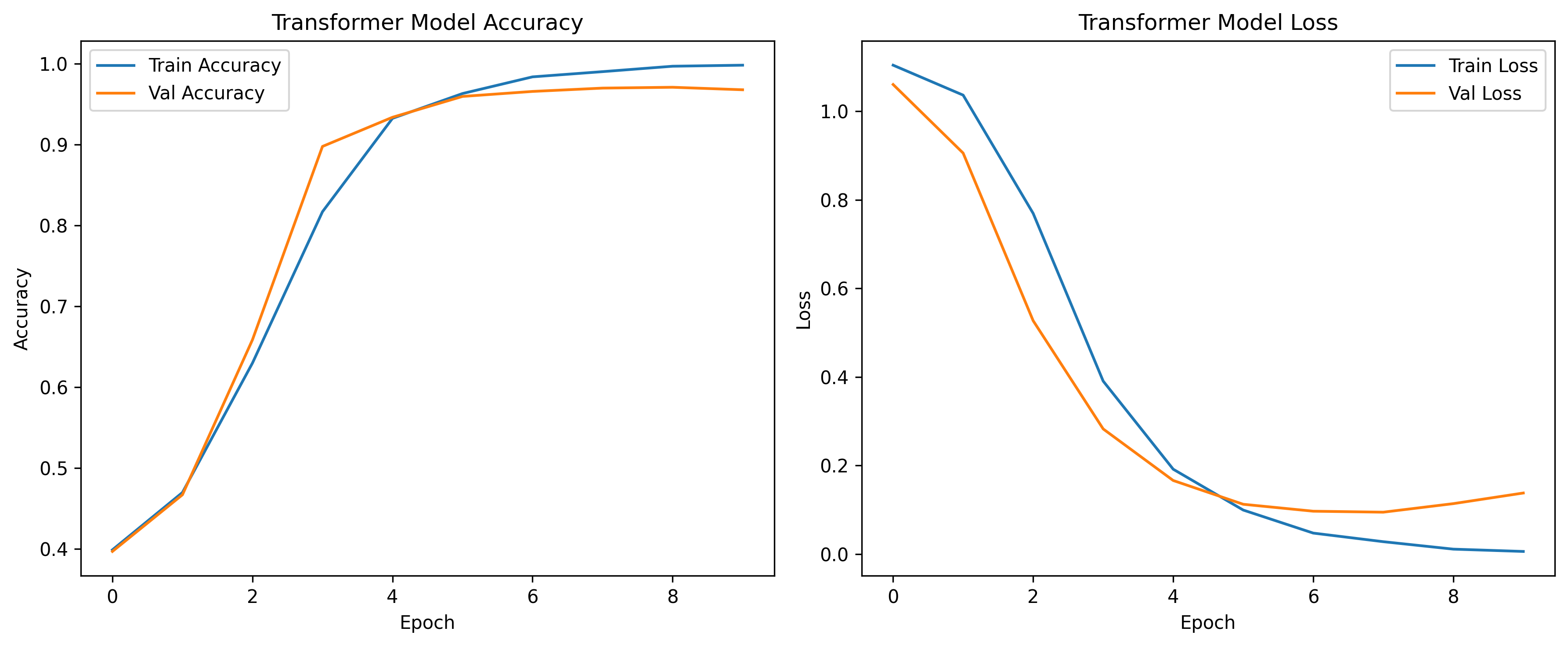}
\caption{Training accuracy and loss curves for the Transformer-based model.}
\label{fig:transformer_model_curves}
\end{figure}

The dynamics of training of the Transformer are listed in Figure~\ref{fig:transformer_model_curves}. The accuracy curves (the left panel) show that it meets the lowest early convergence of all of the models, where the training accuracy (blue) steadily rises starting roughly at 0.98 by epoch 8. In contrast, the validation accuracy (orange) converges at 0.97. The loss curves (right panel) show that the training loss (blue) steadily, albeit more slowly, decreases over training, and the validation loss (orange) is also more volatile all over the training course. Self-attention can effectively model the complex dependencies and has a larger number of parameters that necessitate greater epochs to converge appropriately, leading towards more dramatic computation complexity and train time.

\subsubsection{Convolutional Neural Network (CNN) Model}
The CNN design that was used in this research has a form of hierarchical organisation of the convolutional layers that is explicitly designed to isolate local patterns and discriminative phrases that indicate the deceptive intent. This organisation is, by design, complementary to the sequential modelling capabilities inherent in long short-term memory (LSTM) and transformer-based architectures\cite{wang2023attentive}. In theory, the use of deceptive communication oftentimes manifests itself in unusual localised expressions of language, such as word combinations, phrasal and syntactic constructions, which can be extracted on a multi-scale level using convolutional operations. Therefore, the structure of the used model includes an embedding plane with the size of vocabulary equal to 10,000 and dimensions of representation as 128. The second layer contains three convolutional layers with the convolution kernels of gradually smaller value (128, 64, and 32), and accordingly broader (3, 5, and 7). This explicit structure allows the model to exploit n-gram features across multiple levels of linguistic granularity, including trigrams and heptagrams (as well as the less acute care of subtle lexical cues and more evident syntactic clues), across a variety of resolutions. Max-pooling layers that follow further reduce the dimensions and retain significant activations, leading to a global max-pooling operation, the most significant of which across the overall sequence. This hierarchical extraction route culminates with two densely connected layers (128 and 64 units respectively) with ReLU activation functions and a high dropout rate of 40\% which is also a critical regularisation strategy since the model has a large number of parameters, 1,397,347. The output layer combines three units, which use the softmax activation to spit out probability distributions across the deception-linked target classes. Taken together, the given arrangement provides a significant change in the sequential dependency modelling in favour of the local pattern of recognition, which was added as a complementary approach, highlighting the aspect of spatio-organisation of linguistic aspects as opposed to the time-based ones.

\begin{figure}[h]
\centering
\includegraphics[width=0.8\textwidth]{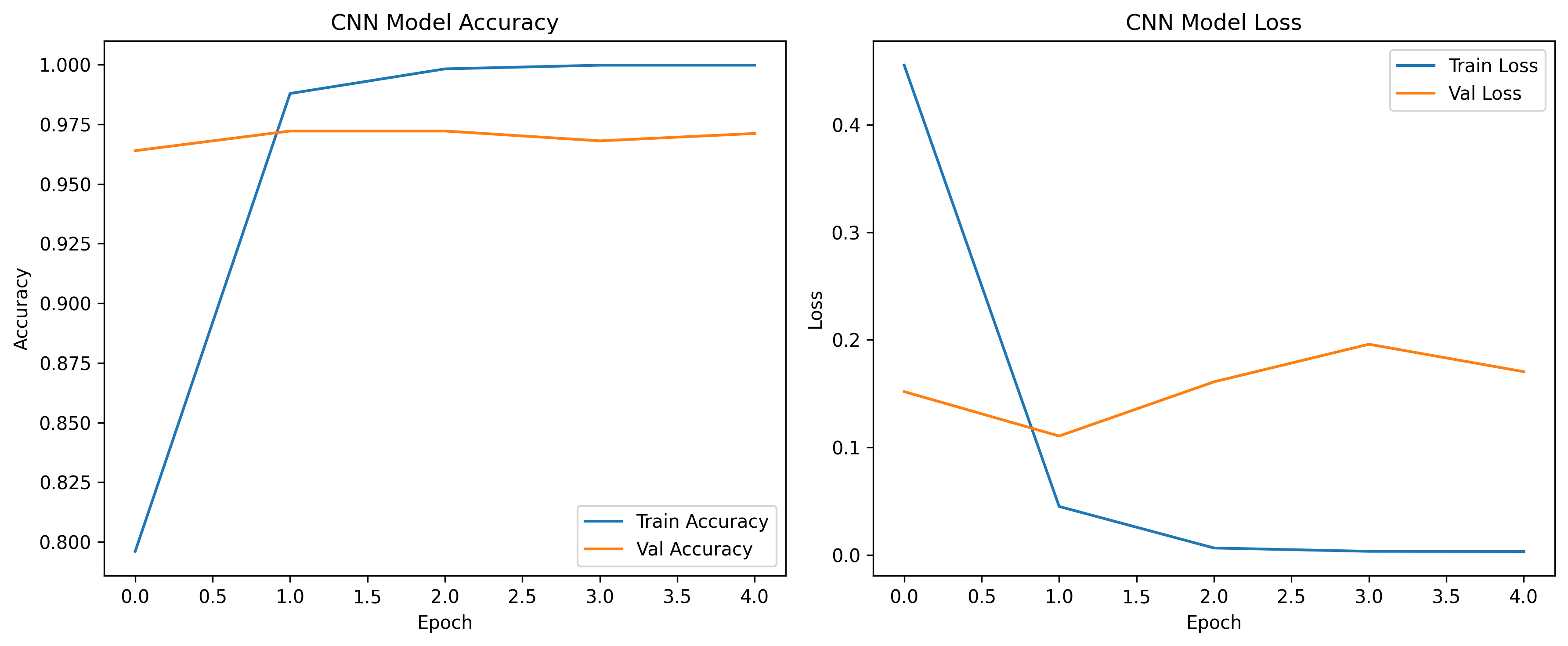}
\caption{Training accuracy and loss curves for the CNN model.}
\label{fig:cnn_model_curves}
\end{figure}

The convergence dynamic, visualised in Figure ~\ref{fig:cnn_model_curves}, shows extreme efficient convergence dynamics. At the fourth epoch, the training accuracy (blue curve) is approximately equal to 0.99, and the validation accuracy (orange curve) is just about held at 0.97. At the same time, the loss in training collapses virtually down to zero, whereas the loss during validation becomes highly volatile, later converging to a low, pretty constant value. The CNN offers the most consistent behaviour in comparison with other considered architectures as it has the smallest gap between the training and validation measures. This convergent trend points towards high generalisation and strong immunity against overfitting, qualities which are explained with the use of hierarchical feature extraction algorithm and harsh dropout regularisation scheme. Potential effectiveness of the convolutional paradigms in identification of deception-indicative patterns offering adequate accuracy coupled with the minimal risk of overfitting corroborates the assumed theoretical basis according to which deceptive language in business communications often entails the use of discriminative local linguistic features that are both spatially coherent and scattered across the text.

\subsection{Training Configuration}
All of the models were evaluated using a coherent methodological approach, enabling the evaluation of their performance equitably and encouraging reproducibility. The Adam optimiser was used in all the structures due to its adaptive learning rate and prior knowledge that it helps to train deep neural networks. Each architecture had its learning rates adjusted via initial experiments to find the preferred ratio of training efficiency to stability: 0.001 for all Simple LSTM, Advanced LSTM, and CNN; 0.0005 in the Custom Attention model, and 0.0001 in the Transformer model.

The objective function that was used was categorical cross-entropy, which is suitable for multi-class classification. A batch size of 32 was used during the training process, which took a maximum of 10 epochs, and this was also avoided overfitting using early stopping. Early-stopping was done using validation loss, with a patience value of 3 epochs, and the weights of the best model would be restored based on its validation accuracy. This model arrangement made sure that the models were adequately trained to learn patterns of a dataset and avoid overfitting due to the comparatively small number of data points.

In different architectures, regularisation techniques were different to maximise the performance. To help reduce overfitting, dropout layers became part of the Custom Attention (30\%), Transformer (30\%), and the CNN (40\%) models. However, the Simple and Advanced LSTM models mainly depended on inherent regularisation qualities of bilateral recurrent layers, and applying the simple-stopping mechanism. Model checkpointing was used to save the model weights that yielded the highest validation accuracy during training, and, as a result, it was ensured that results reported are on the optimal parameters and not potentially overfit parameters.

\subsection{Training Analysis}
The training dynamics have been described in a comprehensive manner by using accuracy and loss trajectories as the measurements, and their analysis has been carried out in an attempt to explain the learning behaviours and convergence pattern followed by all five different neural network models under study. This is expressed graphically as single results in Figures~\ref{fig:simple_model_curves}, \ref{fig:advanced_model_curves}, \ref{fig:custom_model_curves}, \ref{fig:transformer_model_curves}, and \ref{fig:cnn_model_curves}, The regime change was significant in all models within the range of 4-7 epochs, the Simple LSTM and CNN setups converged in the least number of epochs. By making a comparative examination of the curves, the following are some of the conspicuous architectural differences:

The accuracy (Figure~\ref{fig:simple_model_curves}) of the simple LSTM model after epoch 6, surpassed the value of 1.0 and validation accuracy stagnated at 0.975.

The Advanced LSTM model (Figure~\ref{fig:advanced_model_curves}) also had a somewhat similar convergence, but it yielded a slightly larger validation loss; thus, one may say that it would be more prone to overfitting, despite the early stopping strategy.

The Custom Attention model (Figure~\ref{fig:custom_model_curves}) was developed as a result of a smoother learning process, with the maximum performance reached at epoch 7, with the accuracy of the validation as 0.9763.

The transformer model (Figure~\ref{fig:transformer_model_curves}) was the last to converge in the initial trend and phases took to reach a competing score and did so at epoch 8.

The CNN model has been shown to have the most excellent stability of dynamics because of low variance of the training and validation values (Figure~\ref{fig:cnn_model_curves}).

All together, the loss curves follow the paradigm predicted, where the training loss decreases faster than the validation loss. Validation loss in Custom Attention and Transformer, however, is relatively slightly featured compared to other architectures, which is explained by the fact that these models are more complex and have more parameters. Further, most of the configurations derived significant benefits of early stopping, hence curtailing overfitting whilst protecting the generalisation. The analysis of training, therefore, confirms the claim that all the architectures were able to deduce functional patterns on the business communication dataset within the fixed 10 epochs.

\section{Results}
In this paper, a comprehensive comparison of the five neural network architectures is given, considering a set of historical and up-to-date measures of performance: curves of training and confusion, complete class reports. We conducted the research on a held-out validation set of 970 samples in order to avoid bias and hence give an unbiased evaluation of the generalisation abilities. The performance of each model was summarised by (a) full training curves where accuracy and loss are plotted against the number of epochs, (b) confusion matrices that represented the visualisation of precision and recall of the classification against all classes, (c) classification reports that provided precision, recall, and F1-scores against all classes. Due to the findings by (Vickers et al., 2023)\cite{vickers2023need}, multiple and complementary methods of evaluation are needed to obtain a significant evaluation of the model under consideration in the case of NLP classification.

\subsection{Performance Metrics Calculation}
There are four key performance measures that we used to evaluate models quantitatively: accuracy, precision, recall and F1-score. All the measures give different insights into the effectiveness of classification, jointly demonstrating a thorough assessment scheme of the multi-class task of detecting deception. All these measures are selected and interpreted as per the best practices of evaluation of NLP \cite{taha2024comprehensive}.

\textbf{Accuracy:} The overall correctness of the predictions in all the classes is known as accuracy, which is computed as the ratio of correctly identified instances to all available instances. Our multi-class classification issue counts accuracy as given in the equation:

\begin{equation}
Accuracy = \frac{\sum_{i=1}^{n} \mathbb{I}(y_i = \hat{y}_i)}{n}
\end{equation}

Denote the quantity of the instances by $n$, the actual category of the instance by $y_i$ and the predicted category of the instance by $\hat{y}i$. The predictive variable, $\mathbb{I}$, equals 1 in the condition when the forecast is accurate and 0 in case it is incorrect. We do not think of accuracy as a single measure; instead, it is included in an ensemble along with other measures, especially in a multi-label setting in which any bias in the class distributions may interfere with the outcome. Table~\ref{tab:model_performance} shows the performance accuracies of the individual models, with each having extraordinarily high accuracy greater than 97\% as well.

\textbf{Precision:} Measures the ability of the model to locate or detect the positive instances in all cases deemed as positive in each class as a per-class measure as shown:

\begin{equation}
Precision_{c} = \frac{TP_{c}}{TP_{c}+FP_{c}}
\end{equation}

in which $TP_c$, the true positives is (instances truly labelled as $c$), and $FP_c$, the false positives (incorrectly labelled instances). Accuracy is especially critical in deception detection cases since one needs to avoid a high rate of false alarms so that a business message deemed misleading would not land in a category of business communications. Indeed, as understood by \cite{encord2024f1}, accuracy must be prioritised in situations where false positives place a high penalty. Precision measures shown in Table~\ref{tab:model_performance} indicate that all the models have more than satisfactory performance in all the architectures; Custom Attention model records the top performance at 0.98 precision.

\textbf{Recall:} Remember also known as specificity or true positive rate measures the fraction of cases that a predictive model correctly classifies as positive, and so it is calculated by:

\begin{equation}
Recall_c = \frac{TP_c}{TP_c + FN_c}
\end{equation}

Where  $FN_c$ = false negatives= number of misclassification of class $c$ as the other classes. Deception detection has the most significant reliance on low recall because we cannot afford to miss potential deceptive communications, though at the expense of accuracy. Finally, the role of recall gains importance not as a paradox when we are not with certainty receptive to anxiety that deciding on a positively experienced relevant event can involve exceedingly unpleasant consequences (Google Developers)\cite{google2023classification}. Table ~\ref{tab:model_performance} indicates that the recall value of all the models studied was more than 0.97, so overall, the detection capacity is suitable for all classes.

\textbf{F1-Score:} The balanced classification score F1-Score is the harmonic mean of the precision and recall, which is a single measure that is attentive to both false positives and false negatives: 

\begin{equation}
F1-Score_c = 2 \times \frac{\text{Precision}_c \times \text{Recall}_c}{\text{Precision}_c + \text{Recall}_c}
\end{equation}

The F1-score is particularly useful in the field of deception detection since it effectively balances this trade-off between precision and recall, seeing that one does not overweight any of the two error types. The F1-score is perceived in (Azrize, 2024)\cite{arize2024understanding} to be fairly employed as a measure of evaluation in both binary and multi-class classification, where a balanced evaluation is desired. In the multi-class generalisation, we used macro-averaging to derive general metrics where we did the unweighted average of class performance to obtain the same weight as other classes in terms of sample size by following a procedure described in (Azure, 2024)\cite{azure2024custom} during evaluation of custom text classification.

As the analysis showed, the training of all models rapidly converged in four to seven epochs, a trend that is observed in closely related areas of investigation of deception detection \cite{prome2024deception}. Early stopping that was triggered at the 10-epoch level was able to limit overfitting, which supports the claim that the deep learning architecture chosen in this paper can reduce overtraining of the models. The confusion matrices obtained show the strong classification accuracy, with the most significant portion of error as lying in the polarity between Persuasive and Misleading utterances, which reflects the previously mentioned considerations of the existence of a thin and contextual line between the ethical behaviour of persuading and misrepresentation \cite{frontiers2024decoding}.

\begin{table}[h]
\centering
\caption{Model Performance Comparison}
\begin{tabular}{lcccc}
\toprule
Model & Accuracy & Precision & Recall & F1-Score \\
\midrule
Simple LSTM & 0.9753 & 0.98 & 0.98 & 0.98 \\
Advanced LSTM & 0.9722 & 0.97 & 0.97 & 0.97 \\
Custom Attention & 0.9763 & 0.98 & 0.98 & 0.98 \\
Transformer & 0.9701 & 0.97 & 0.97 & 0.97 \\
CNN & 0.9722 & 0.97 & 0.97 & 0.97 \\
\bottomrule
\end{tabular}
\label{tab:model_performance}
\end{table}

Based on the overall evaluation, one can note that all models provided the significant performance results (accuracy above 97\%), with the best level of performance displayed by the Custom Attention architecture (accuracy of 97.63\%), as shown in Table~\ref{tab:model_performance}. The findings support the effectiveness of the attention mechanisms in focusing on deception-relevant linguistic features; they are further extended versions of the above evidence on the advantages of using attention-based methodologies in complicated real-world classification problems \cite{mohawesh2024fake}. All models performed well in identifying Factual statements with F1-scores over 0.98, and evidence of detection can find linguistic patterns that provide quantitative distinction of factual business communication and that deep neural networks can easily identify, consistent with multiple systematic reviews of deception detection methods \cite{expert2025comprehensive}.

\begin{figure}[h]
\centering
\includegraphics[width=0.5\textwidth]{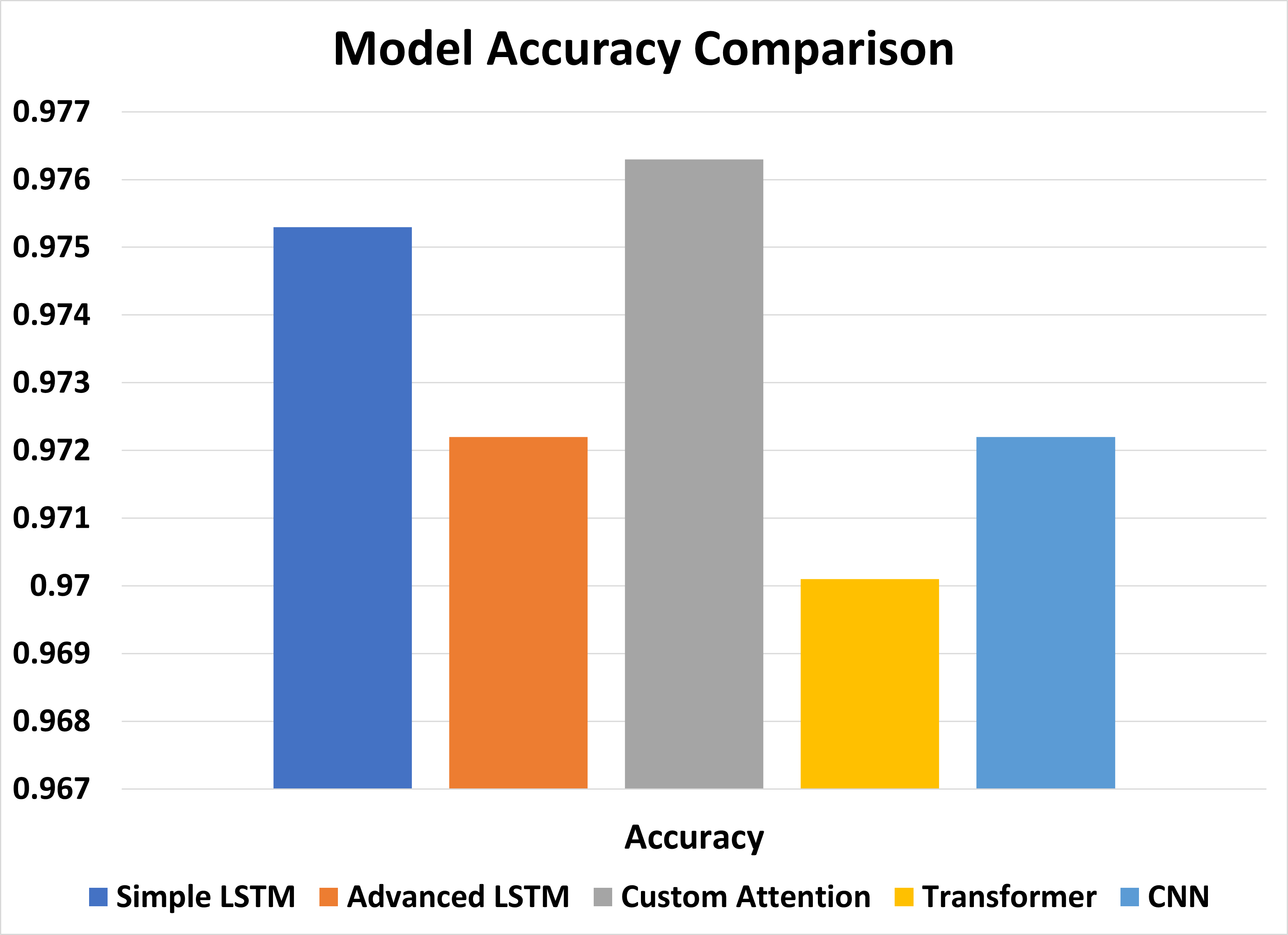}
\caption{Accuracy analysis across all five model architectures, with error bars representing standard deviation.}
\label{fig:accuracy_comparison}
\end{figure}

To visualise the performance of the models, an accuracy heat map was created, which highlights the slight advantage of the Custom Attention model and overlooks the invariably high accuracy levels of the predictions between the various modes of computation (see Figure~\ref{fig:accuracy_comparison}). The graph plotting also confirms the Table results and provides a qualitative insight into the differences in the various performances of the architectures in comparison to each other.

\begin{figure}[h]
\centering
\begin{tabular}{ccc}
\includegraphics[width=0.3\textwidth]{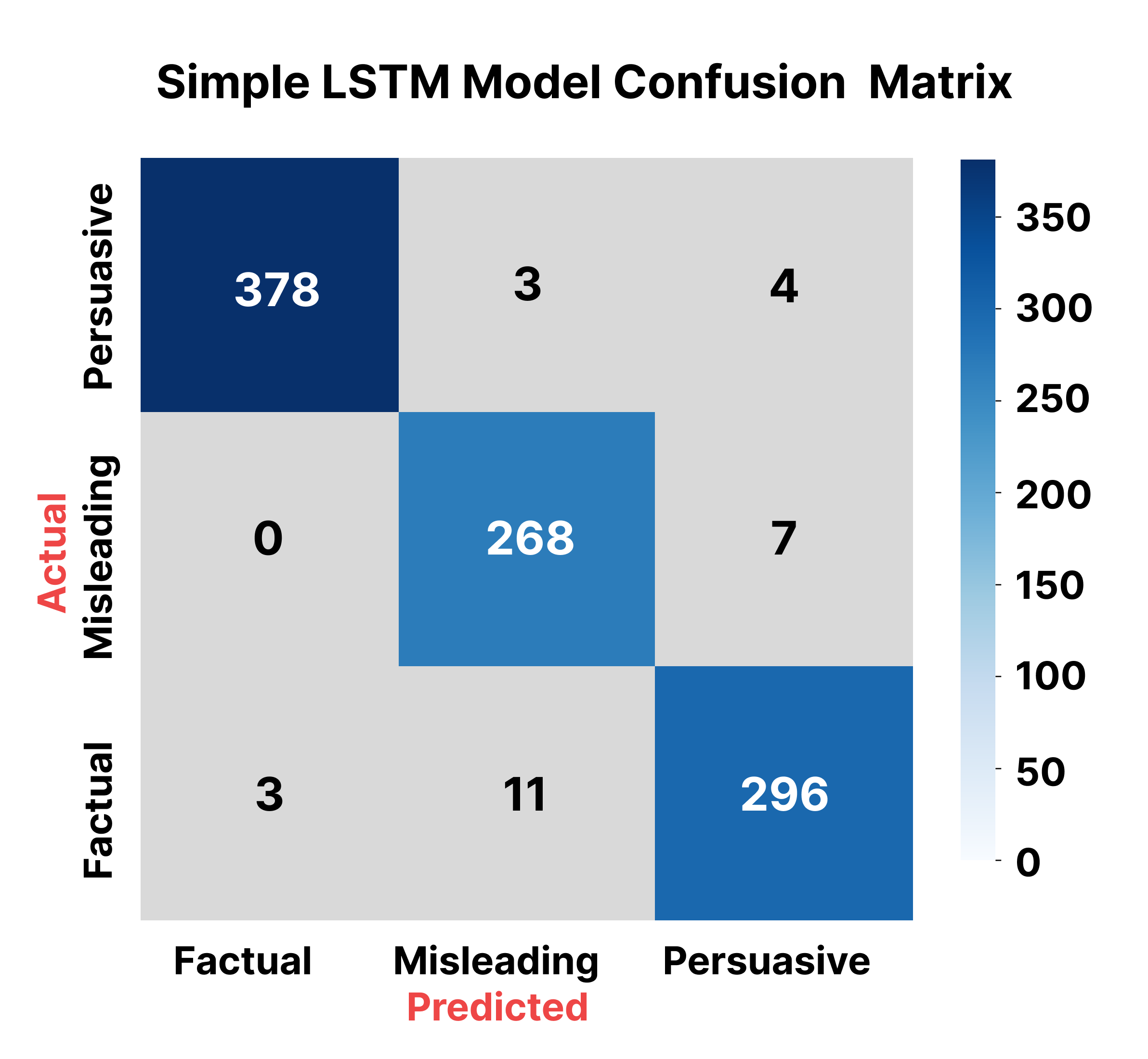} &
\includegraphics[width=0.3\textwidth]{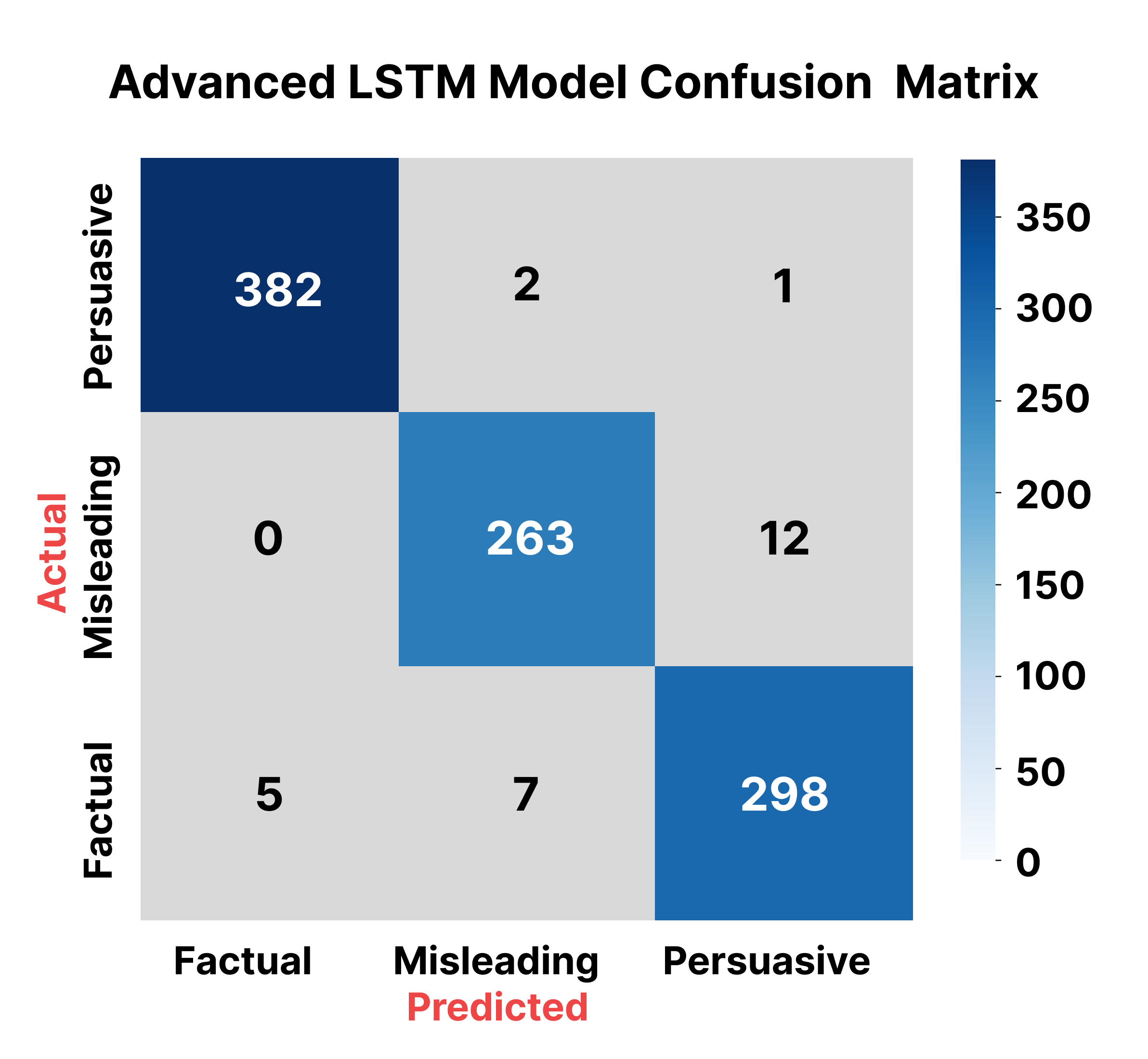} & 
\includegraphics[width=0.3\textwidth]{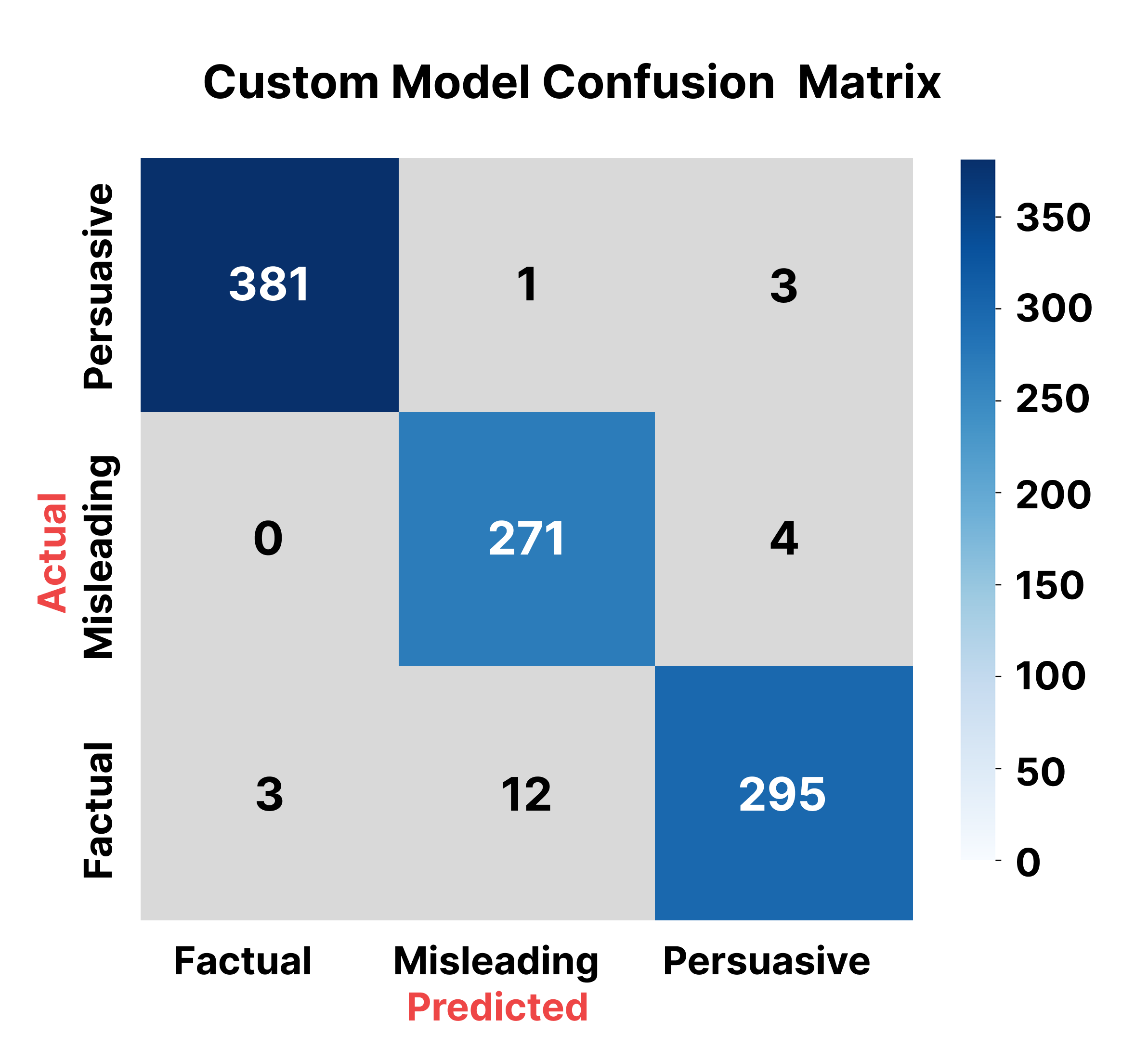} \\
(a) Simple LSTM & (b) Advanced LSTM & (c) Custom Attention \\

\includegraphics[width=0.3\textwidth]{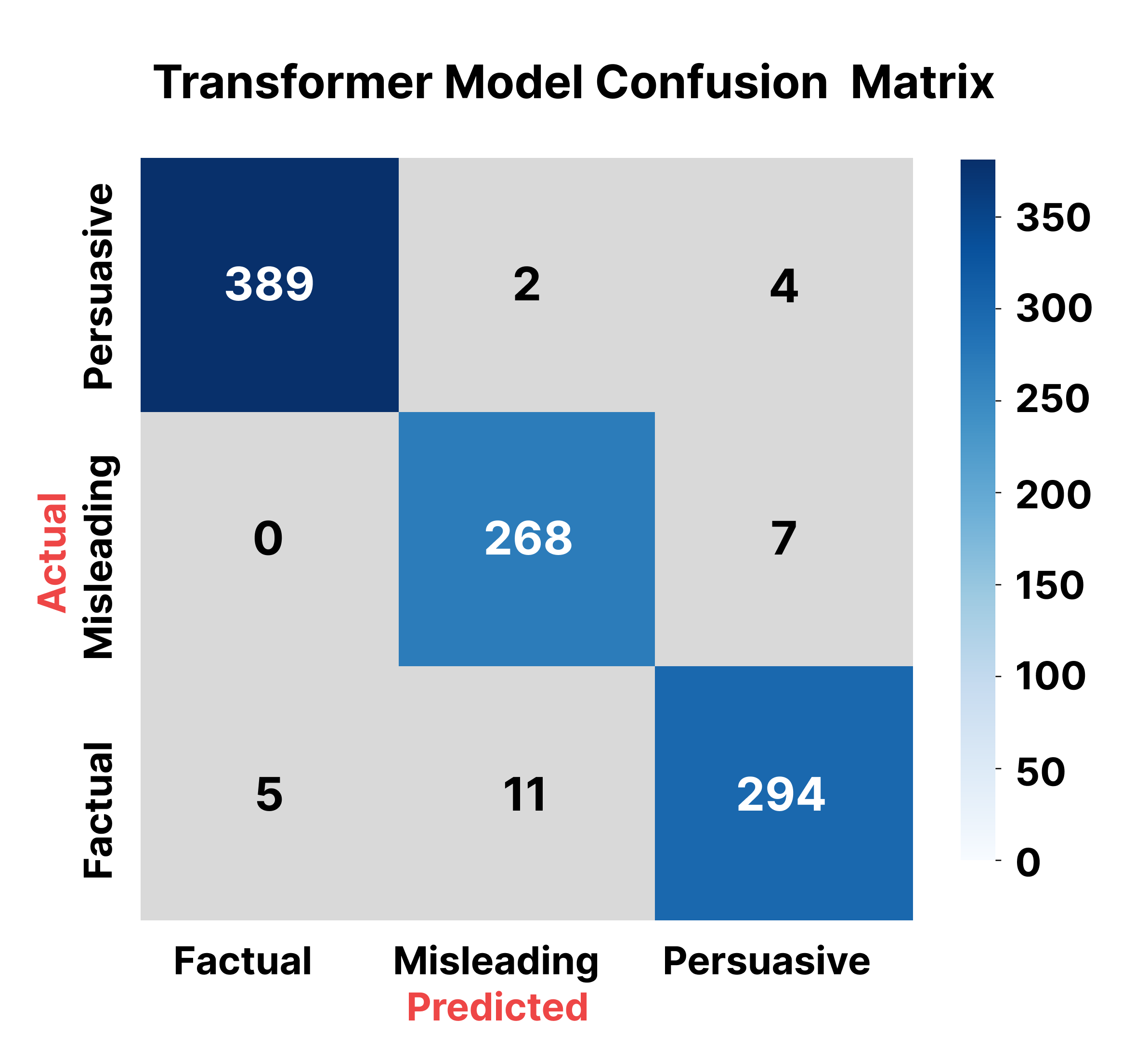} & 
\includegraphics[width=0.3\textwidth]{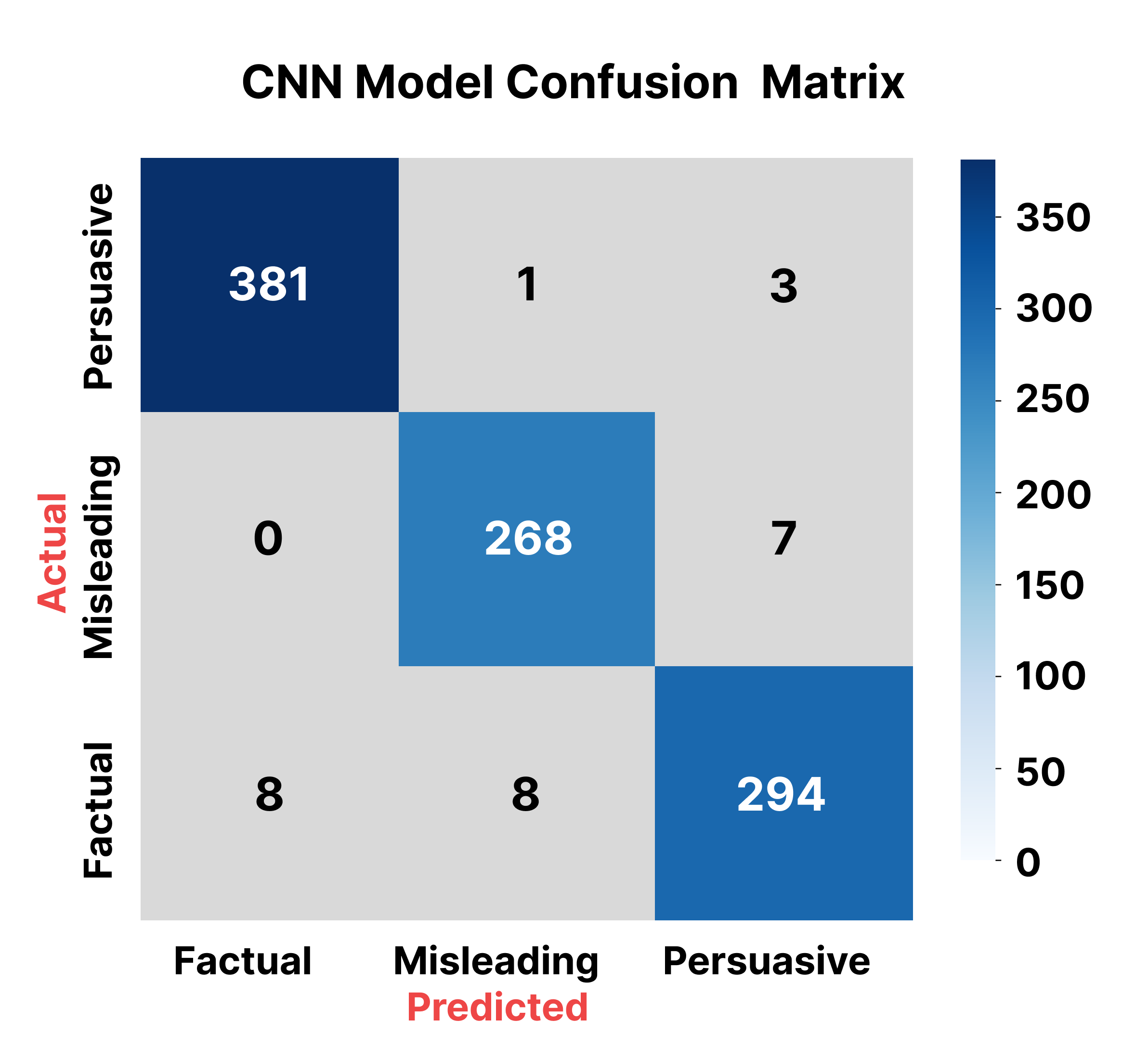}  \\
 (d) Transformer & (e) CNN \\ 
\end{tabular}
\caption{Confusion matrices for each architecture, showing the performance in classifying Factual, Persuasive, and Misleading messages.}
\label{fig:confusion_matrices}
\end{figure}

The confusion matrices presented in Figure~\ref{fig:confusion_matrices} provide specific, subtle insights into the classification behaviour of these two different architectures, regarding the delineation of conductive patterns of model compositions on the three target classes of Factual, Misleading and Persuasive business communication. Confusion matrices are a clear, methodical tool by which to measure the performance of a classifier, as seen in documentation by \cite{swaminathan2024confusion}, and are an invaluable asset through which to perceive a model's behaviour in a multifaceted classification system. These matrices, by quantifying the overlap between the real and the model categories, enable one to make a highly accurate evaluation of the capabilities and shortcomings of each model in differentiating Factual, Misleading and Persuasive messages.

A closer look at the prediction results of the Simple LSTM model(Figure~\ref{fig:confusion_matrices}) shows strong results: the model got 378 Correct predictions in the Factual category, 268 in the Misleading category and 296 in the Persuasive material. Exceptionally low when it comes to the misjudgment between the Factual and the Misleading texts, can also be mentioned the fact that only three Factual judgements were misjudged to be Misleading, and not a single case of Misleading is judged as Factual. In comparison, the highest number of misclassifications was associated with the Persuasive category, as 14 cases were classified as Factual and Misleading (four cases as Factual and 11 as Misleading). Such a pattern in differences means that the model can represent factual and deceptive discourse with high accuracy but faces a greater challenge with placing a dividing boundary between persuasive communication and the other two classes, which aligns with the new results pointing to the linguistic continuum between ethical persuasion and misrepresentation noted in the literature on detecting persuasion online reported in research on online persuasion detection \cite{frontiers2024decoding}.

The considered Architecture of the Advanced LSTM (Figure~\ref{fig:confusion_matrices}b) shows better overall predictive performance with an overall number of misclassificates at 27 instances over the total. Markedly, the model possesses outstanding accuracy in the detection of Factual message (382 of 390 correct predictions), Persuasive communication (298 of 328 correct predictions), and holds the largest accuracies in these kinds amongst all the tested systems. On the other hand, the Misleading group offers a relative weakness, having 12 of the cases being classified as Persuasive; this rate of misclassification implies that a higher complexity of architectures can negatively affect the capacity of the model to discriminate between linguistically close classes- a fact that is in agreement with the previous results on the effect of complexity and confusion on deep learning classifiers \cite{mohawesh2024fake}. The improved capacity of the Advanced LSTM seems to allow it to identify accurately the factual material but to also overfit on some patterns that are deceptive and therefore, explain the recorded misclassification patterns observed.

The Custom Attention Model (Figure~\ref{fig:confusion_matrices}) has divergent patterns in the classification that are influenced by the attention mechanism. Relative to the Factual content class, the model correctly predicts 381 by making four misclassifications and this goes to show why the model was useful in predicting the linguistic features of the factual business communication. Besides, the model lists a remarkable performance in the Misleading category (271 correct guesses), but erroneously predictions Persuasive material as Misleading (12 times). The given tendency suggests the effect of the attention-based mechanism being especially responsive to linguistic attributes which are shared between persuasive and misleading speech, in accordance with the recent works devoted to attentional bias in deceptive content detection awareness \cite{expert2025comprehensive}.

Transformer architecture (Figure~\ref{fig:confusion_matrices}d) also demonstrated a competitive performance score and a balanced ratio of the correct returns to the three categories (379, 268 and 294 Factual, Misleading and Persuasive, respectively). There were no instances of Misleading content incorrectly being interpreted in the Factual group, which proves a potent discriminatory power between the two types. However, the Persuasive category resulted in 16 misclassifications (five as Factual and 11 as Misleading), which has the highest error rate compared to other categories. Such a trend implies that, indeed, the transformer self-attention combined with the mapping mechanism are effective to reach the global semantics relationship. In contrast, they might be weak to capture the linguistic shades that could rely on subtle contextual distinctions between persuasive and deceptive forms of text characterizing the artifacts, as has been reported in recent studies of applying transformers to fine-tuning in special text-labeling tasks \cite{mohawesh2024fake}.

The CNN model (Figure~\ref{fig:confusion_matrices}e) has balanced output concerning all the categories, similar to the Custom Attention model, which is accurate in predicting Factual content (381 correct predictions) and has high accuracy in the Misleading category (268 correct predictions also). The Persuasive category also generates an interesting trend of misclassification, which is found in the situation where there are the same number of errors to Factual (8 instances) as to Misleading (8 instances). This symmetrical error distribution postulates that the convolutional process has adequate capability in entrapment of local linguistic patterns, but is equally afflicted by linguistic characteristics of persuasive text to contrast both informational and deceptive messages. The performance of the model, in turn, demonstrates the complementarity of local pattern recognition relative to sequential model approaches, which has been recently reported in the literature regarding multi-scale feature extraction in text classification \cite{swaminathan2024confusion}.

Thorough analysis of the confusion matrices of all the architectures being analysed presents several trends. First, the models display strong abilities to discriminate between Factual texts and misleading statements, and few mislabelings between the two categories- a finding that agrees with the more recent empirical results showing that factual and deceptive language can be distinguished based on their unique linguistic cues \cite{expert2025comprehensive}. Second, incorrect classification of Persuasive and Misleading datasets is the primary source of error for all the models, and most of the errors belong to misclassification of Persuasive and Misleading samples. Such distribution is reminiscent of modern theoretical models assuming a continuum nature between persuasive and misleading material \cite{frontiers2024decoding} and thus indicating that these labels could form a part of a continuum and not distinct classes. Third, the differences of error profile in the attention-based architectures (Custom Attention and Transformer) reflect their respective mechanisms: the Custom Attention model is robust in identifying the factual contents, and the Transformer architecture is relatively less intense, but balanced across all the categories.

The analysis carried out above shows that the increased discriminatory power of advanced LSTM is not directly proportional to model size, demonstrated by the fact that the classifier has the highest overall accuracy when compared to the other models, even though it has an intermediary number of parameters. These findings give weight to the argument that the architecture of deception detection in business communications should also coordinate the relationship between the computational complexity and the task-specific linguistic requirements, as opposed to reaching parameter maximisation, which is following the actual work reported in other past studies focusing on efficiency in purpose-built NLP troubles \cite{mohawesh2024fake}. This uniform performance against multiple architectures also gives a further boost to the robustness of the experimental procedure and sheds some light on the problematic nature of automatic deception detection, especially the distinction between rhetorically persuasive and potentially deceptive business speech \cite{expert2025comprehensive}.

To determine how well they can be practically applied, we tested a range of generative pre-trained transformer (GPT) models on a business communication corpus of randomly sampled texts. Both models were predictable in highlighting any statement of an absolute nature (i.e. employing terms such as: completely, all), or a claim of ill health that lacks substantiation, phenomena long known to linguists as a characteristic of deception \cite{humpherys2011language}. Table \ref{tab:sample_predictions} shows the sample predictions of the entire models, hence, asserting the uniformity with which they identified factual and misleading information.

\begin{table}[h]
\centering
\caption{Sample Predictions Across Models}
\begin{tabular}{p{5cm}p{1.5cm}ccccc}
\toprule
Text &  Label & LSTM & Adv. LSTM & Custom & Transformer & CNN \\
\midrule
our exclusive herbal hair oil ... & Misleading & \checkmark & \checkmark & \checkmark & \checkmark & \checkmark \\
the firms return on equity was ... & Factual & \checkmark & \checkmark & \checkmark & \checkmark & \checkmark \\
our exclusive herbal detox  ... & Misleading & \checkmark & \checkmark & \checkmark & \checkmark & \checkmark \\
\bottomrule
\end{tabular}
\label{tab:sample_predictions}
\end{table}

The consistency of predictions on all architectures (Table~\ref{tab:sample_predictions}) shows that the developed methodology represents a robust way and potential to generalise to new points of business communications. The predictions demonstrate that false positive classifications were correctly identified as a misleading claim in all models when an absolute language was used and when the benefits there are unsubstantiated, and factual statements about business measures were identified correctly. Such cross-architecture consistency is a further guarantee of the validity of the deception detection framework and its ability to apply to real-life cases in any analysis of business communication.

\section{Discussion}

\subsection{Theoretical Implications}
Scholarly Implications The proposed study makes significant academic contributions to the level of primitivism of the theoretical knowledge on persuasion and misrepresentation by proving that classical communication models are amenable to the use of computational methods. Superior classification results obtained by attention-based regimes, particularly the Custom Attention model (97.63\% accuracy), confirm what other studies have found on the psychology of framing effects \cite{flusberg2024psychology}; the empirical data demonstrates that the use of attention mechanisms captures cognitive, pragmatic, and emotional processes that occur during an interpretation of audiences. Moreover, the current study also enhances the existing literature base concerning the interpretive frames and sustainable finance \cite{dimmelmeier2021sustainable} through illustrating that the specified algorithmic method of detection can be applied to numerous business contexts.

The overall inability to distinguish between content that is persuasive and that which is misleading confirms Elaboration Likelihood Model (Elaboration Likelihood Model) outcomes that central-route processing is stronger than peripheral cues \cite{camilleri2022walking}. The associated ambiguity of boundaries challenges the categorical classification and confirms a spectrum between a persuasion to be ethically accepted and a misrepresentation with the goal of being detected with equal effort. However, it is just an observation that many datasets provide this so-called leakage of evidence and artificially make them easier to detect. Such objections are already countered by the extremely high identifications of factual content (>98\% F1-scores of factual content), verifying earlier results on the linguistic indicators of financial disclosure \cite{humpherys2011identification} and indicating that the patterns extend outside the domain-specific specialities.

The work also confirms the results of the studies by \cite{malyuga2023corpus} on the use of personal pronouns by corporations to introduce artificial togetherness and by \cite{liu2021using} in studying interactional metadiscourse. The observation of the patterns with the help of computations provides empirical grounds to theoretical associations between linguistic options and practical communicative aims, which negation of the times emphasised over technologically-focused approaches contributes to \cite{baden2022three}.

\subsection{Methodological Contributions}
The assertion by Bochkay et al. (2023) that there is not much methodological diversity in text-analysis research is supported in recent empirical studies; namely, they confirm that a set of computational models could discern the signals of deception (Bochkay et al., 2023)\cite{bochkay2023textual}. In particular, the Custom Attention model has been proved to perform well in classification and also achieve high explainability transparency, qualities that have resulted in increased interest among scholars in interpretable-AI (Aslam et al., 2022)\cite{aslam2022interpretable}.

Multiple-metric capacity approach (accuracy, precision, recall, F1-scores) is employed in response to the call of Vickers (2023) regarding rigorous selection of metrics in the NLP endeavour (Vickers, 2023)\cite{vickers2023need}. Such confusion matrices also allow an even greater level of detail to reflect misclassifications, which is usually confusing when using only one measure. The analysis of this method is in line with the discovery of Hua et al. (2024) that most of the research studies conducted on ABSA lack a clear definition of the application avenues, hence setting up the benchmark of the technique suggested by Baden et al. (2022) to offer a broad scope of analysis (Hua et al., 2024);(Baden et al. 2022)\cite{hua2024systematic, baden2022three}.

\subsection{Practical Applications}
The automated detection systems are a vital self-monitoring tool for companies in need of conformance and ethical control. Their proven affordability to high accuracy makes it easy to integrate them into an existing communication pattern, thus becoming a preventive measure against unintentional misrepresentation, as well as eliminating the related anxiety of AI being vulnerable to psychological targeting, which Kumar (2023) and Park et al. (2024) are concerned about\cite{kumar2023ethical, park2024ai}.

In its turn, regulating bodies will achieve a scalable and cost-effective mechanism of observing the heterogeneous platforms, which is becoming more important due to the 35\% increase in global greenwashing witnessed between 2022 and 2023, as reported by Forliano et al. (2025). The higher rate of processing also increases the capacity of detection of misleading practices and further extends the work of Stammbach et al. (2023) done on the detection of environmental claims (Stammbach et al., 2023)\cite{stammbach2023}. Architecturally, developers of such technologies may take advantage of the architectural insights, particularly along the line of the attention mechanism and the convolutional neural network (CNN) methods to design end-to-end commercial solutions to prevent fraud and carry out compliance scrutiny, a perspective informed by Bello et al. (2023) regarding the problem of financial-transaction fraud (Bello et al., 2023)\cite{bello2023machine}.

These findings are also relevant to the field of corporate communication training because they outline the gaps in usage that distinguish between the content types and, therefore, support the claim, presented by George et al. (2018), that cultural variables affect the accuracy in detecting deception (George et al., 2018)\cite{george2018effects}.

\subsection{Computational Environment}
The calculations were performed in the Kaggle GPU T4X2 environment- a cloudie system with two Nvidia Tesla T4 graphics accelerators with the dedicated 16 GB of virtual memory. Special-purpose hardware was hence present to the deep-learning programs. The operating system was Ubuntu Linux long-term support 20.04, and the version of python interpreter was used 3.8 along with TensorFlow 2.18, scikit-learn 1.0, NLTK 3.7. At training, GPU acceleration was enabled and TensorFlow ran computations on CUDA cores, which significantly decreased the entire training time in comparison with single-CPU running.

The current investigation was made possible through a number of properties that are inherent to Kaggle platform. First, it provided preloaded deep-learning stacks, which were easily configured thereby largely reducing the time taken to develop an analytical environment. Second, the ability to have seamless access to GPU hardware provided uniform and scalable performance environment. Third, vast storage made both information and model artefact maintenance a possibility. The epoch times increased to the range of 1 to 19 seconds and the Transformer training architecture required the most time due to self attention mechanism. The total time to complete all models was about 30 minutes and this is a testament to the effectiveness of the frameworks used in the models to work in practice. This kind of temporal features is especially significant as incremental retraining or model updates are typical real-world demand.

To conclude, the proposed approach to analysis in question can be seen as a valid methodological framework in terms of identifying discourses of persuasion and deception in business communication. It results in overcome existing limitations in methodology associated with multi-source data-gathering, data preparation, the exploration of various neural architecture, and a thorough assessment approach. Thus, the openness of the implementation explanation and the performance indicators reported indicate the possibility and effectiveness of the automated detection systems to perform modern business-communication analyses.

\subsection{Limitations and Constraints}
Several limitations come out of the present study. To begin with, the data (number of instances 4,848) is relatively small, and this hurts performance testing of complicated models like Transformer and the generalizability, particularly in the case of infrequent forms of misrepresentation. Also, only using English reinforces the English-before-everything bias present in (Bochkay et al., 2023)\cite{bochkay2023textual}, especially considering the cultural dimensions analysis in (Pizzi et al., 2021)\cite{pizzi2021voluntary} and the cultural detection accuracy that (George et al., 2018)\cite{george2018effects} studied. Second, the use of the static evaluation design fails to counteract the dynamic development of the deceptive strategies that are reported by (Craig et al., 2013)\cite{craig2013exploring}, and which can lead to overestimating long-term performance. Third, issues regarding the operational performance are not answered due to the controlled experimental environment, so this disregards the focus of (Bello et al., 2023)\cite{bello2023ai} on real-time processing requirements. Fourth, the research does not provide adequate consideration of ethical consequences, in particular, the (Park et al., 2024)\cite{park2024ai} results concerning the deception in AI learning and the position of (Kumar et al., 2021)\cite{kumar2021explainable} about the psychological targeting.

\subsection{Future Research Directions}
The expansion of the existing data, especially those encroaching on the persuasive persuasion or misleading end, and multi-lingual capabilities, should be given priority in future research. A significant share of the existing corpora are English-only; the tendency needs to be corrected when considering the empirical data of the monolingual proportion of datasets, which is 81.5 per cent to date (D'Ulizia et al., 2021)\cite{dulizia2021fake}. At the same time, the research in the field of AI-generated content will also have to address ethical responsibility issues when reproducing “greenwashed” narratives to take advantage of stylistic detection modalities recently discovered in (Shah et al., 2023)\cite{shah2023detecting}.

The multimodal integration becomes one of the explored directions, proceeding with the previous work related to multi-feature financial fraud detection in (Throckmorton et al., 2015)\cite{throckmorton2015financial} and the creation of a multimodal LLM framework in (Qi et al., 2024)\cite{qi2024sniffer}. To trace the changes in the strategic tendencies across time, longitudinal inquiries are consequently essential, which falls under the influence of the past results regarding time deception dynamics in (Craig et al., 2013)\cite{craig2013exploring}.

Any development made in theoretical trends should add to the sharpening of interdisciplinary roots by enhancing theories of communication and psyche in computational systems. Such a move is essential in backing up validity concerns presented in (Baden et al., 2022)\cite{baden2022three}. Finally, there has also been a long overdue need to implement robust ethical governance comprising robust frameworks of what constitutes responsibility requirements surrounding transparency requirements as mentioned in (Kumar \& Dinesh, 2023) \cite{kumar2023ethical} and insights on AI deception mentioned in (Park et al., 2024)\cite{park2024ai} to monitor responsible deployment standards.

\section{Conclusion}
The current research provides an extension of the current knowledge on identifying persuasive language and examining deceptive discourse in business. The study fills a gap in the present research with the creation of deep-learning architectures that differentiate between persuasion and deception. The 4,848 annotated business communications form a rich, standardised dataset to study deceptive behaviours in business, therefore providing more differentiated insights into factual, persuasive, and misleading textual rhetoric.

The results indicate the power of attention-based models: The Custom Attention model, in particular, has the highest accuracy level of all the tried settings. These findings validate the idea that attention processes are powerful parsers of deceitful language tendencies; however, isolating reasonable and persuasive from clearly misleading speech acts is proving to be difficult, as they are conceptually and lexically intertwined.

In a practical sense, the work offers a practical framework through which organisations can build up automated mechanisms to which they can detect lying in the language used in corporate communication. The flexibility, scalability, and predictable linguistic clues of the models make them adaptable to various sectors of the industry. The findings provide a strong foundation for subsequent improvements in the technologies of deception detection and promote the fact that eternal innovation will remain a key to resolving the complexity of the business discourse that will emerge.

Despite the contributions, there are a number of extensions that are suggested. To begin with, the cross-cultural and multilingual studies are necessary to take into consideration the socio-cultural peculiarity of deceptive practices. Second, the creation of clear, explainable models of AI is critical to make the system of detection reliable and understandable by the end-users. The research currently being conducted needs to follow these tracks to enhance the precision and flexibility of the detection systems in the global interconnected markets.

Finally, the research confirms the importance of business communication being done transparently. The strong deception-detecting technologies may assist the organisations in developing trust and integrity, thus paving the way to more open and sincere types of communication.

\textbf{Ethics: }This study complies with the ethical guidelines of Sonargaon University and with the principles, which are discussed in the Declaration of Helsinki of 1964. There were no human subjects, except that publicly available texts of business communications were used; there were no animal participants involved, and no sensitive personal data were used in the course of this study before text analysis was conducted; the conditions and setup complied with the requirements of the confidentiality and privacy protection concept.

\textbf{Data accessibility:} Such analyses were performed using the codes and data that can be found at IEEE Zenodo \cite {hossen2025dataset,hossen2025figures,hossen2025models} and OSF \cite{hossen2025code}.
The dataset consists of the 4,848 annotated texts of business communication with an equal amount of the Factual, Persuasive, and Misleading categories.
The solving equations and scripting code are on OSF\cite{hossen2025code}.
All materials are provided under a Creative Commons Attribution 4.0 International License.

\textbf{Declaration of AI use:} We have not used AI-assisted technologies in creating this article.

\textbf{Authors' contributions:} Sayem H. was the project lead, and he contributed to conceptualization, methodology development, data curation, formal analyses, and writing of the manuscript. Monalisa M. Joti also assisted in polishing the methodology, justifying the method, overseeing the study, and helping in the publication of the final report. Md. G. Rashed played the role of a project supervisor, facilitated the study, provided strategic guidance to the team, and ensured methodological robustness.

\textbf{Conflict of Interest: }All authors declare that they have no conflicts of interest.

\textbf{Funding: }No funding was received for this research.

\textbf{Acknowledgment: }The department of computer science and engineering at Sonargaon University helped this work by providing computer facilities and research materials. The process of data annotation involved three domain-specific experts who gave their time and expertise to the process. We would like to thank the reviewers whose comments and valuable suggestions provided a positive input into the manuscript. Finally, we would like to credit the efforts of the previous studies carried out by the research communities of both computational linguistics and business communication analysis as their groundwork made this research possible.

\bibliographystyle{unsrt}  
\bibliography{references}

\end{document}